\documentclass[lettersize,journal]{IEEEtran}
\usepackage{amsmath,amsfonts}
\usepackage{booktabs}
\usepackage{wrapfig}
\usepackage{algorithmic}
\usepackage{array}
\usepackage[caption=false,font=normalsize,labelfont=sf,textfont=sf]{subfig}
\usepackage{textcomp}
\usepackage{stfloats}
\usepackage{url}
\usepackage{verbatim}
\usepackage{graphicx}
\usepackage{balance}
\usepackage{xcolor}         
\usepackage{amsmath}
\usepackage{gensymb}
\usepackage{subfig}
\usepackage{subcaption}
\usepackage{adjustbox}
\usepackage{tabularx}
\usepackage{makecell}
\usepackage{multirow}
\usepackage{hhline}
\usepackage{makecell}
\usepackage{array}
\usepackage{comment}
\usepackage{enumitem}
\usepackage[dvipsnames]{xcolor}
\usepackage[table]{xcolor}
\usepackage{arydshln}
\usepackage{resizegather}
\usepackage[pagebackref=true,breaklinks=true,letterpaper=true,colorlinks,bookmarks=false]{hyperref}
\usepackage{mathtools}
\usepackage{orcidlink}

\definecolor{firstcolor}{RGB}{255, 205, 210}
\definecolor{secondcolor}{RGB}{227, 242, 253}

\definecolor{resred}{RGB}{240,128,128}
\definecolor{resblue}{RGB}{135,206,250}

\newcommand{\dovored}[1]{\textcolor{resred}{#1}}
\newcommand{\openvblue}[1]{\textcolor{resblue}{#1}}

\newcommand{\changed}[1]{\textcolor{black}{#1}}

\newcommand{\tableholder}{%
  \colorbox{white}{\textcolor{white}{00}-\textcolor{white}{00}}/%
  \colorbox{white}{\textcolor{white}{00}-\textcolor{white}{00}}/%
  \colorbox{white}{\textcolor{white}{00}-\textcolor{white}{00}}%
}

\newcommand{\tableholdern}{%
  \colorbox{white}{\textcolor{white}{0}-\textcolor{white}{0.}}/%
  \colorbox{white}{\textcolor{white}{0.}-\textcolor{white}{.0}}/%
  \colorbox{white}{\textcolor{white}{.0}-\textcolor{white}{0}}%
}

\begin{document}
\title{360DVO: Deep Visual Odometry for Monocular 360-Degree Camera}
\author{
Xiaopeng Guo\,\orcidlink{0009-0004-4982-5457}, 
Yinzhe Xu\,\orcidlink{0009-0001-7459-3828}, 
Huajian Huang\,\orcidlink{0000-0002-0963-1146}, \textit{Member, IEEE},
and Sai-Kit Yeung\,\orcidlink{0000-0001-7974-0607}, \textit{Senior Member, IEEE}
\vspace{-0.8cm}
\thanks{
This work was supported by the HKUST Marine Robotics and Blue Economy Technology Grant, and the Marine Conservation Enhancement Fund (MCEF20107 and MCEF22112). \textit{(Corresponding authors: Huajian Huang.)}

Xiaopeng Guo and Yinzhe Xu are with the Division of Integrative Systems and Design, Hong Kong University of Science and Technology (e-mail: xguoay@connect.ust.hk; yxuck@connect.ust.hk).

Huajian Huang is with the Department of Computer Science and Engineering, Hong Kong University of Science and Technology (e-mail: hhuangbg@connect.ust.hk).

Sai-Kit Yeung is with the Division of Integrative Systems and Design, the Department of Computer Science and Engineering, and the Department of Ocean Science, Hong Kong University of Science and Technology (e-mail: saikit@ust.hk).
}}


\maketitle

\begin{abstract}
Monocular omnidirectional visual odometry (OVO) systems leverage 360-degree cameras to overcome field-of-view limitations of perspective VO systems. However, existing methods, reliant on handcrafted features or photometric objectives, often lack robustness in challenging scenarios, such as aggressive motion and varying illumination. To address this, we present 360DVO, the first deep learning-based OVO framework. Our approach introduces a distortion-aware spherical feature extractor (DAS-Feat) that adaptively learns distortion-resistant features from 360-degree images. These sparse feature patches are then used to establish constraints for effective pose estimation within a novel omnidirectional differentiable bundle adjustment (ODBA) module. To facilitate evaluation in realistic settings, we also contribute a new real-world OVO benchmark. Extensive experiments on this benchmark and public synthetic datasets (TartanAir V2 and 360VO) demonstrate that 360DVO surpasses state-of-the-art baselines (including 360VO and OpenVSLAM), improving robustness by 50\% and accuracy by 37.5\%. Homepage: \textit{\url{https://chris1004336379.github.io/360DVO-homepage}}

\end{abstract}

\begin{IEEEkeywords}
visual odometry, omnidirectional vision
\vspace{-0.2cm}
\end{IEEEkeywords}

\setlength{\fboxsep}{1.2pt}

\section{Introduction}\label{sec:intro}

\IEEEPARstart{V}{isual} odometry (VO) and simultaneous localization and mapping (VSLAM) estimate agent's ego-motion from image sequences which enable various applications, including autonomous navigation and augmented reality. 
360-degree cameras capture omnidirectional field-of-view (FoV) information and produce full-sphere images in the widely used equirectangular projection, thereby providing substantially richer scene coverage than perspective sensors. 
Usage of 360-degree cameras in VO system has emerged as a practical and effective solution to enhance system performance. Early works, such as OpenVSLAM~\cite{openvslam} and 360VO~\cite{360VO}, extend typical VO systems~\cite {orb, dso} to exploit the merit of the 360-degree camera. However, without revisiting feature representations and overall pipeline, existing omnidirectional visual odometry (OVO) systems remain suboptimal in real-world challenges characterized by significant lighting variations, low frame rates, motion blur, and so on.
\IEEEpubidadjcol
\begin{figure}[t]
    \centering
    \includegraphics[width=\linewidth]{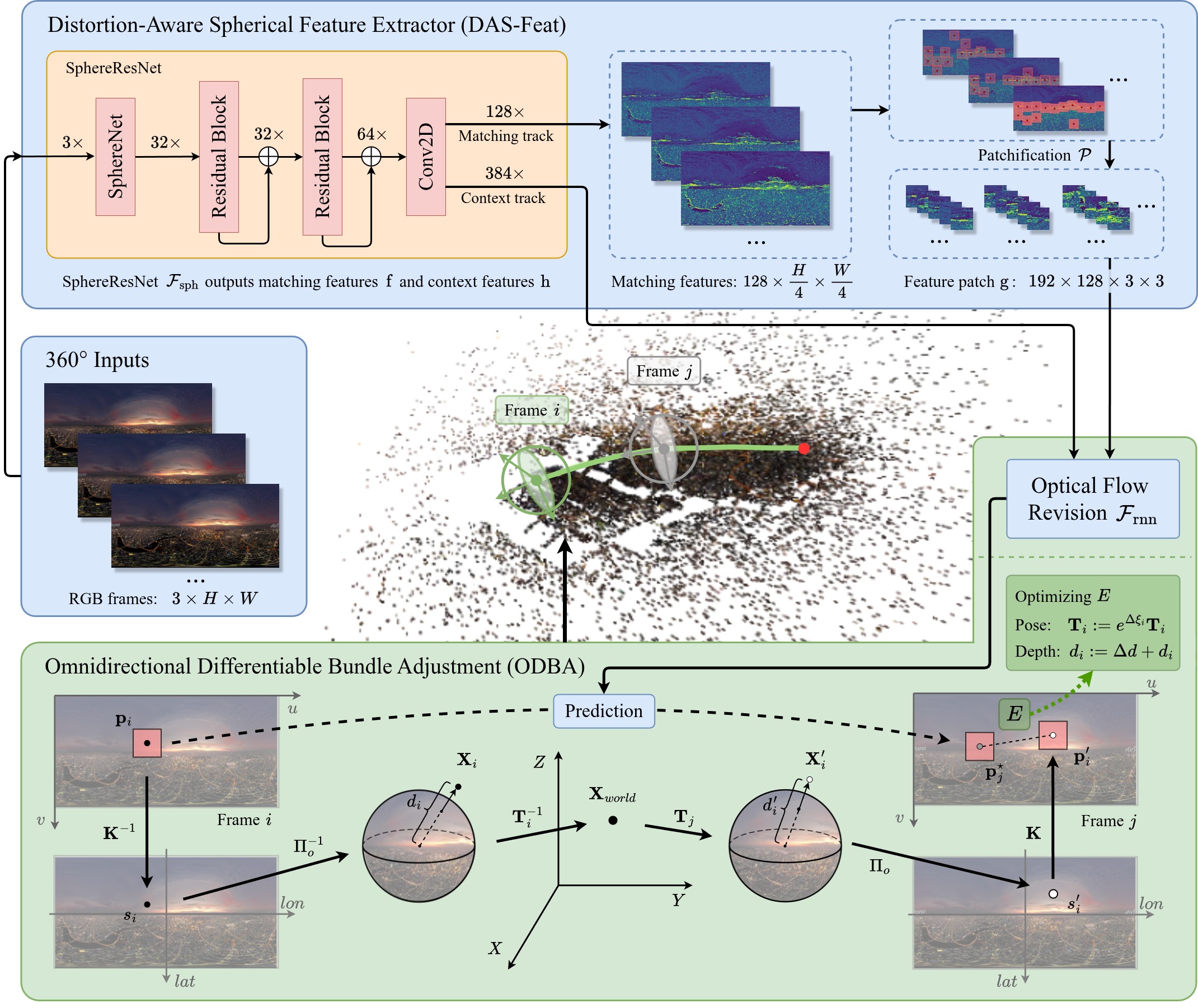}
    \caption{Framework overview of 360DVO. 
    Our method takes sequential 360-degree RGB frames as input and extracts matching features and context features using our proposed DAS-Feat module (Sec.~\ref{sec:sphresnet}) on each of them. 
    In DAS-Feat, the key component SphereResNet extracts distortion-resistant features, allowing patches to be cropped without deformation. 
    After patchifying (Sec.~\ref{sec:patch}) the matching features around their gradient maxima, we compute the correlation of patch features and context features and estimate optical flow through a recurrent network. In the ODBA module, the pose $\mathbf{T}_i$ and depth $\mathbf{d}_i$ of frame $i$ are jointly optimized by minimizing the distance between predicted patch $\mathbf{p}^{\star}_j$ (from optical flow) and reprojected patch $\mathbf{p}_i'$ on an adjacent frame $j$ (Sec.~\ref{sec:optim}).
    }
    \vspace{-0.5cm}
    \label{fig:fw}
\end{figure}

In this work, we aim to enhance OVO performance in challenging scenarios by leveraging deep learning features. 
However, one of the prominent properties of 360-degree images is the strong nonlinear distortion caused by projection.
Most deep feature extractors~\cite{superpoint,dino, DroidSLAM,DPVO,dinov2,gim} trained on large‑scale perspective images implicitly assume linear sampling, which produces unreliable features from omnidirectional images, degrading the stability and accuracy of pose estimation. 
Additionally, computational power would be wasted in the non-linear region, reducing the running speed. 
Therefore, feature extraction with distortion perception capabilities and efficient omnidirectional pose optimization constraints are the key elements for building robust and precise deep learning-based OVO systems. 

To narrow the gap, we propose 360DVO, a novel deep omnidirectional visual odometry system using a monocular 360-degree camera. 
We develop a distortion-aware spherical feature extractor (DAS-Feat) with a novel spherical residual network SphereResNet to extract robust features from omnidirectional images.
The SphereResNet integrates SphereNet~\cite{SphereNet} with residual blocks~\cite{resnet}, which allows learning and accommodating projection‑induced distortions effectively.
Instead of using dense feature maps, we formulate the pose‑estimation problem over sparse feature patches, similar to DPVO~\cite{DPVO}. 
Furthermore, we derive an omnidirectional differentiable bundle adjustment (ODBA) component suitable to jointly optimize omnidirectional camera poses and depth of 3D points. These components exploit the unique properties of spherical imagery and improve accuracy and efficiency of our system. The overview of the 360DVO framework is illustrated in Fig.~\ref{fig:fw}.

Recognizing the critical need for real-world evaluation, we present a new benchmark dataset comprising 360 videos filmed in 20 different real-world scenes. Finally, we conduct experiments on both real-world and synthetic benchmark datasets to comprehensively verify the efficacy of our approach. Our 360DVO achieves state‑of‑the‑art performance than other baselines even in challenging scenarios.

In summary, the contributions of this work include:

\begin{itemize}

    \item We propose 360DVO, the first deep omnidirectional visual odometry framework that learns ego-motion from monocular 360-degree videos.

    \item We develop a distortion-aware spherical feature extractor (DAS-Feat) built on a novel network SphereResNet to obtain reliable omnidirectional image features.

    \item We derive deep spherical feature constraints and establish omnidirectional differentiable bundle adjustment (ODBA), enabling efficient optimization of camera poses.
    
    \item We present a new challenging benchmark dataset dedicated to comprehensively evaluate OVO methodologies performance in real-world scenarios. 
\end{itemize}

\vspace{-0.2cm}
\section{Related Work}\label{sec:related}
\noindent{\textbf{Omnidirectional Visual Odometry.}} 
Omnidirectional cameras can obtain rich image information and sufficient environmental texture details~\cite{widefov}, which makes it possible to collect the necessary data and bridge the research gaps.
Prior works~\cite{omni_lsdslam, omnidso, ORBSLAM3} expand the field of view by employing fisheye cameras and adapting indirect and direct VO/SLAM pipelines~\cite{lsdslam, dso, orbslam2} with appropriate projection models~\cite{unify_camera_model, fisheye_camera_model}. Although these methods broaden the field of view, they remain extensions tailored to fisheye cameras, not fully accommodating omnidirectional scenarios. To achieve a 360-degree field-of-view, ROVO~\cite{rovo} builds a wide-baseline multi-fisheye camera system. In contrast to multi-camera setups, OpenVSLAM~\cite{openvslam} supports omnidirectional input captured by a portable 360-degree camera, offering great flexibility.
360VO~\cite{360VO} proposes a direct OVO and is able to recover semi-dense point clouds while having a higher requirement of the input image quality. Some works, such as 360-VIO~\cite{360vio} and LF-VISLAM~\cite{360vio}, integrate IMU into the OVO framework for robustness improvement. 

\noindent{\textbf{Spherical Feature Extraction.}} Traditional Convolutional Neural Networks (CNNs) ~\cite{cnn} typically process 2D images using a rectangular grid structure. For 360-degree imagery, the most common representation is the equirectangular projection, which suffers from significant distortions. Spherical CNNs~\cite{spherical_cnn} encode rotation equivariance into the CNNs for classification. Flat2Sphere~\cite{spherical_cnn} adjusts the kernel size based on spherical coordinates to approximate distortion. EquiConvs~\cite{cfl} and SphereNet~\cite{SphereNet} both perform sampling on the sphere and project the convolution kernels onto the equirectangular image. EquiConvs~\cite{cfl} samples on the sphere and defines the kernel by its angular size and resolution, while SphereNet~\cite{SphereNet} samples and defines the square kernel on the tangent plane. SphereNet~\cite{SphereNet} facilitates fast spherical feature extraction and, due to the structure similarity to traditional CNNs, can be readily integrated into other network architectures. We adapt SphereNet~\cite{SphereNet} with residual blocks and propose SphereResNet as the feature extraction network of 360DVO, which can effectively learn and extract robust,  distortion-resistant features.

\noindent{\textbf{Bundle Adjustment.}} In traditional VO and SLAM systems, bundle adjustment (BA) plays a crucial role in refining camera poses and 3D structure by minimizing reprojection error~\cite{ORBSLAM3} or photometric error~\cite{dso} across image sequences. 
Classical BA frameworks like g2o~\cite{g2o} and Ceres Solver~\cite{ceres} optimize nonlinear cost functions using second-order, gradient-based methods, typically Gauss–Newton or Levenberg–Marquardt.
Some methods have combined traditional BA with deep learning techniques. DeepSFM~\cite{deepsfm} uses two cost volumes to iteratively optimize both camera pose and depth. DROID-SLAM~\cite{DroidSLAM} embeds a differentiable dense bundle adjustment layer and integrates it into the network architecture, enabling end-to-end training, while DPVO~\cite{DPVO} implements a sparse version of it. 
With the recent advances in differentiable rendering, SC-OmniGS~\cite{huang2025scomnigs} has explored a new framework that bundle-adjusting omnidirectional camera poses along with radiance field represented by 3D Gaussians~\cite{kerbl20233dgs}. However, SC-OmniGS is primarily designed for scene reconstruction rather than online VO, and it faces challenges in large-scale settings when initial poses are not provided by an SfM or VO frontend. 
In contrast, we focus specifically on the VO problem and propose an ODBA module that efficiently estimates camera trajectories without reliance on external initialization. 

\section{Methodology: 360DVO}\label{sec:method}
360DVO is composed of two core components, including 
distortion-aware spherical feature extractor (DAS-Feat) and omnidirectional differentiable bundle adjustment (ODBA), as shown in Fig.~\ref{fig:fw}. 
In DAS-Feat, we employ SphereResNet to extract distortion-resistant features from each input omnidirectional image $I_i$ of size $H\times W$.  
\changed{
The convolution kernel size is dynamically adjusted to accommodate nonlinear image projection distortion. 
Benefiting from the distortion-aware features, we can sample unwarped square patches from the extracted feature maps.
Following DPVO~\cite{DPVO}, a sparse patch graph is maintained to encode the patch-to-frame relationships.}
After predicting and revising sparse patch motions, ODBA incorporates spherical reprojection constraints and jointly estimates accurate camera poses and 3D points.

\noindent{\changed{\textbf{Spherical Camera Model Basic.}}}\label{sec:spherical_model}
\changed{
To handle omnidirectional imagery, we adopt a unified spherical camera model~\cite{360VO}, which maps each image pixel to a point on the unit sphere. As illustrated in Fig.~\ref{fig:fw}, the equirectangular image coordinate system is parameterized by longitude angles $\theta \in [-\pi, \pi]$ and latitude angles $\phi \in [-\pi/2, \pi/2]$.
Given $\mathbf{X}=(x,y,z,1)^{T}$ representing the 3D point, the projection $\Pi$ and inverse projection $\Pi^{-1}$ of the spherical camera model are formulated as: 
\begin{equation}\label{eq:sphericla_project}
    \Pi(\mathbf{X})=\left[\begin{smallmatrix}
        u\\v\\1
    \end{smallmatrix}\right]=\mathbf{K}\left[\begin{smallmatrix}
        \theta\\\phi\\1
    \end{smallmatrix}\right]=\mathbf{K}\left[\begin{smallmatrix}
        \arctan(x/z)\\-\arcsin(d\cdot y)\\1
    \end{smallmatrix}\right],
\end{equation}
\begin{equation}\label{eq:sphericla_reproject}
    \Pi^{-1}(\mathbf{p},\mathbf{d})=\left[\begin{smallmatrix}
        x\\y\\z\\1
    \end{smallmatrix}\right]=\frac{1}{\mathbf{d}}\left[\begin{smallmatrix}
        \cos(\phi)\sin(\theta)\\-\sin(\phi)\\\cos(\phi)\cos(\theta)\\\mathbf{d}
    \end{smallmatrix}\right],
\end{equation}
\begin{equation}\label{eq:intrinsic}
    \mathbf{K}
    =\left[\begin{smallmatrix}
        W/2\pi&0&W/2\\0&-H/\pi&H/2\\0&0&1
    \end{smallmatrix}\right],\quad
    \left[\begin{smallmatrix}
        \theta\\\phi\\1
    \end{smallmatrix}\right]=\mathbf{K}^{-1}\left[\begin{smallmatrix}
        u\\v\\1
    \end{smallmatrix}\right].
\end{equation}
where $d=1/\sqrt{x^{2}+y^{2}+z^{2}}$ is the inverse distance from $\mathbf{X}$ to the sphere center, and $\mathbf{d}$ denotes the depth variable associated with the 2D patch.
The intrinsic matrix $\mathbf{K}$ of the spherical model depends solely on the omnidirectional image resolution $(H, W)$ and converts between spherical coordinates $(\theta, \phi)$ and pixel coordinates $(u, v)$.
}

\begin{figure}[t]
    \centering
    \captionsetup[subfloat]{labelfont={rm,footnotesize},textfont=footnotesize}
    \includegraphics[width=\linewidth]{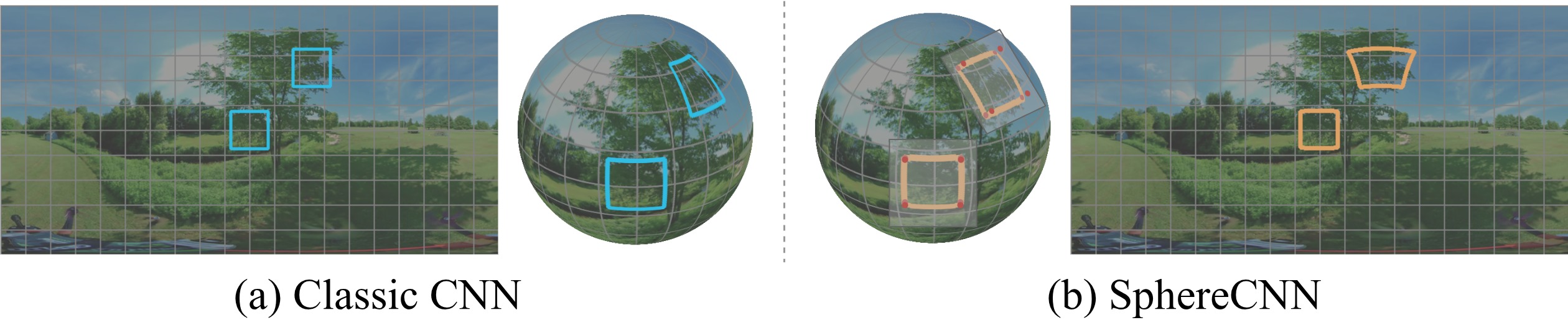}
    \caption{
    Comparison of feature extraction between the classic CNN and the SphereCNN \cite{SphereNet}.
    The distortion-aware convolution kernels differ in their pixel sampling manners along the height in an omnidirectional image, which is guided by the tangent projection across latitudes on the image's corresponding sphere.
    }
    \label{fig:sph}
    \vspace{-0.3cm}
\end{figure}

\vspace{-0.2cm}
\subsection{Distortion-Aware Spherical Feature Extractor}\label{sec:sphresnet}

\noindent{\textbf{SphereResNet.}} 
To address distortion in omnidirectional images, we introduce a network structure coupling a spherical convolution module~\cite{SphereNet} with residual blocks~\cite{resnet}, referred to SphereResNet. 
Each input $I_i$ is fed into the spherical convolution module that inversely projects the omnidirectional image onto a unit sphere. 
A $7\times7$ spherical convolution kernel samples pixels on the sphere's tangent plane and then projects those samples back onto the omnidirectional image.
Fig.~\ref{fig:sph} presents the difference in the pixel sampling between the classic CNN and the spherical convolution module (SphereCNN) \cite{SphereNet}.
Following the spherical convolution module, two pairs of residual blocks with dimensions 32 and 64 are employed to mitigate vanishing gradients and enhance overall performance. The feature map channels are increased to 128 (for matching track) and 384 (for context track) via $1\times1$ convolutions. 
As shown in Fig.~\ref{fig:fw}, the SphereResNet $\mathcal{F}_{\text{sph}}$ ultimately produces matching features $\mathrm{f}_i$ and context features $\mathrm{h}_i$ from the matching and context tracks, respectively:
\begin{equation}
   \mathrm{f}_i, \mathrm{h}_i\xleftarrow{} \mathcal{F}_{\text{sph}}(I_i).
\end{equation}
Distortion‑aware features are obtained by deforming standard convolution kernels, which correct spherical distortions across latitudes and enable more accurate trajectory prediction. 

\noindent{\changed{\textbf{Patchification.}}}\label{sec:patch}  
Leveraging distortion‑resistant features from SphereResNet, square patches are extracted directly from the matching features $\mathrm{f}_i$ without warping via the patchification module $\mathcal{P}$. Assuming constant depth within each patch, patch $k$ in frame $i$ is represented by the homogeneous coordinate $\mathbf{p}_{ki}=(u,v,1)^{T}$ of its center pixel with depth $\mathbf{d}_k$, where $\mathbf{d}_k$ is initialized to the mean depth across patches in last keyframe.
For every patch, $\mathcal{P}$ outputs patchified matching features that are indicated by $\mathrm{g}$, along with coordinates $\mathbf{p}$ and depth $\mathbf{d}$:
\begin{equation}\label{eq:patchify}
    \{(\mathbf{p}_k,\mathbf{d}_k,\mathrm{g}_k)\}_{k=1}^{N}=\mathcal{P}( \mathrm{f}_i),
\end{equation} 
where $N$ is the number of patches extracted per frame. 
Specifically, per‑channel spatial gradients are computed on the matching feature maps and aggregated across channels to obtain a single gradient‑magnitude map. Patch centers are selected as the pixels with maximal values on this map, and a fixed $3\times3$ patch is extracted around each center.

We maintain a sparse patch graph $\mathcal{E}$ whose edges connect patches and frames. 
For a patch $k$ extracted from frame $i$, add edges $\{(k,j)\in \mathcal{E} \mid \vert j-i\vert < r\}$, where $j$ indexes adjacent frames and $r$ is a temporal radius.

\vspace{-0.2cm}
\subsection{Omnidirectional Differentiable Bundle Adjustment}\label{sec:odba}
\noindent{\textbf{\changed{Spherical Reprojection Constraints.}}} 
Based on the spherical camera model defined at the beginning of Sec.~\ref{sec:spherical_model}, we formulate the reprojection relationships used in ODBA.
Let $\mathbf{T}_n \in \textbf{SE}(3)$ denote the camera pose at frame $n$. 
For a patch $\mathbf{p}_k$ observed in frame $i$ with estimated depth $\mathbf{d}_k$, its reprojection into another frame $j$ is computed as:
\begin{equation}\label{eq:reprojection}
    \mathbf{p}_{kj}' = \Pi \left( \mathbf{T}_{ij} \cdot \Pi^{-1} (\mathbf{p}_{ki}, \mathbf{d}_k) \right) , 
\end{equation}
where $\mathbf{T}_{ij}=\mathbf{T}_j\circ\mathbf{T}_i^{-1}$ represents the camera pose transformation from frame $i$ to $j$. 
\changed{This operation constrains the reprojected spherical coordinates across frames, allowing ODBA to jointly refine the camera poses and geometry.}

\noindent{\textbf{\changed{Optical Flow Revision.}}}
\changed{
For each edge $(k,j)\in \mathcal{E}$, the correlation volume $\langle\,\mathrm{g}_k, \mathrm{h}_j\rangle$ is computed between the patchified matching features $\mathrm{g}_k$ from $\mathcal{P}$ and context features $\mathrm{h}_j$ from $\mathcal{F}_{\text{sph}}$. Specifically, $\mathbf{p}_k$ is reprojected from frame $i$ to $j$ by Eq.~\ref{eq:reprojection}. A square feature map centered at the reprojection is cropped from $\mathrm{h}_j$, and inner products with $\mathrm{g}_k$ are computed. Then the update operator $\mathcal{F}_{\text{rnn}}$~\cite{DPVO}, a recurrent network, takes $\langle\,\mathrm{g}_k, \mathrm{h}_j\rangle$ as input and estimates the 2D motion (optical flow) of patch $k$ from frame $i$ to frame $j$: 
\begin{equation}            
\mathbf{p}_{kj}^{\star}=\mathbf{p}_{ki}+\mathcal{F}_{\text{rnn}}\bigl(\langle\,\mathrm{g}_k, \mathrm{h}_j\rangle \bigl),
\end{equation}
where $\mathbf{p}_{kj}^{\star}$ is the predicted center point of patch $k$ in frame $j$.
}

\noindent{\textbf{\changed{Nonlinear Optimization.}}}\label{sec:optim}
\changed{ODBA jointly optimizes camera poses and 3D point depths by minimizing the coordinate error between the predicted and reprojected patches. 
The resulting nonlinear problem is solved using the Gauss–Newton algorithm by minimizing the following cost function $E$:
\begin{equation}\label{eq:optim}
\underset{\mathbf{T}_i,\mathbf{T}_j,\mathbf{d}}{\textrm{arg\,min}}\enspace E\,\,=\sum_{(k,j)\in\mathcal{E}} \left\| \mathbf{p}^\star_{kj} - \mathbf{p}_{kj}' \right\|^2_{\sum_{kj}}.
\end{equation}
Using Lie algebra and the chain rule, the Jacobians of the reprojection patch $\mathbf{p}'$ with respect to $\mathbf{T}_i$, $\mathbf{T}_j$, and $\mathbf{d}$ are computed, namely $J_i,\, J_j\in\mathbb{R}^{2\times6}$ and $J_d\in\mathbb{R}^{2\times1}$. Detailed derivations are provided in the supplementary material. The corresponding Hessian matrix is then constructed from these Jacobians, and Eq.~\ref{eq:optim} is solved by: 
\begin{equation}\label{eq:hessian}
\left[ 
\scalebox{0.85}{$\displaystyle
\begin{array}{cc:c}
\mathbf{w}J_i^TJ_i&\mathbf{w}J_i^TJ_j&\mathbf{w}J_i^TJ_d\\
\mathbf{w}J_j^TJ_i&\mathbf{w}J_j^TJ_j&\mathbf{w}J_j^TJ_d\\ \hdashline
\mathbf{w}J_d^TJ_i&\mathbf{w}J_d^TJ_j&\mathbf{w}J_d^TJ_d
\end{array}
$}
\right]
\left[\begin{smallmatrix}\Delta\xi_i\\\Delta\xi_j\\\Delta \mathbf{d}
\end{smallmatrix}\right]
=\mathbf{e}^{\star}
\left[\begin{smallmatrix}
\mathbf{w}J_i^T\\\mathbf{w}J_j^T\\\mathbf{w}J_d^T
\end{smallmatrix}\right],
\end{equation} 
where $\xi_i,\,\xi_j\in \mathfrak{se}(3)$ represent the Lie algebra coordinates of $\mathbf{T}_i$ and $\mathbf{T}_j$, $\mathbf{e}^{\star}=\left|\mathbf{p^{\star}}-\mathbf{p}'\right|$ represents the reprojection error and $\mathbf{w}$ is the matrix of weights, obtained from $\mathcal{F}_{\text{rnn}}$. }

The Hessian matrix can be blocked and solved for the updates to camera poses $\Delta\xi$ and depth $\Delta\mathbf{d}$ by Schur complement: \begin{equation}
    \begin{bmatrix}
\mathbf{A}_{[12\times12]}&\mathbf{B}_{[12\times1]}\\\mathbf{B}^T_{[1\times12]}&\mathbf{C}_{[1\times1]}
    \end{bmatrix}\begin{bmatrix}
        \Delta\mathbf{\xi}_{[12\times1]}\\\Delta \mathbf{d}_{[1\times1]}
    \end{bmatrix}=\begin{bmatrix}
        \mathbf{r}_{ij}\\\mathbf{r}_d
    \end{bmatrix}.
\end{equation}

\changed{The obtained updates are finally applied to optimize the camera poses by $\mathbf{T}\coloneqq \exp^{\Delta\xi}\mathbf{T}$ and depth by $\mathbf{d}\coloneqq\Delta\mathbf{d}+\mathbf{d}$.}

\begin{figure*}[t]
    \centering
    \includegraphics[width=\linewidth]{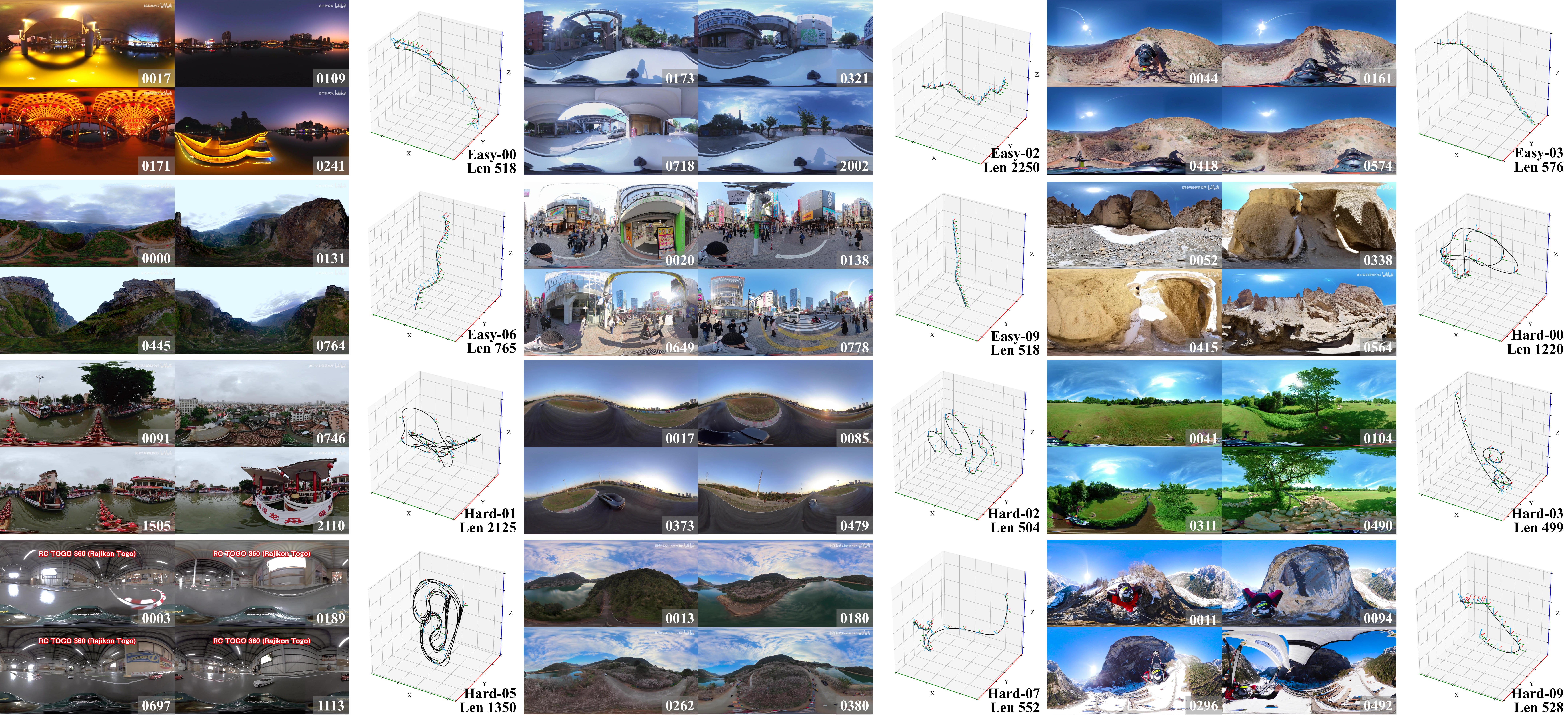}
    \caption{Sample sequences of 360DVO dataset, demonstrating representative frames, 3D trajectory, and length for each sequence.
    }
    \vspace{-0.3cm}
    \label{fig:dataset}
\end{figure*}

\section{360DVO Dataset}\label{sec:dataset}
Existing datasets for evaluating OVO methods remain limited in scope and realism. 
360VO~\cite{360VO} offers 10 synthetic urban sequences of omnidirectional images with smooth trajectories. Recently, TartanAirV2~\cite{tartanairV2} further expands a large-scale simulation SLAM dataset TartanAir~\cite{tartanair} by introducing additional environments and modalities, such as fisheye and panoramic views. In addition, it provides multi-modal sensor data and precise ground truth labels, including stereo RGB images, depth maps, segmentation, optical flow, and camera poses. Though pioneering in OVO evaluation, both datasets lack real-world complexity, limiting their utility for benchmarking robustness in practical applications.

To address this gap, we present a large-scale real-world OVO benchmark dataset, partially illustrated in Fig.~\ref{fig:dataset}. 
Our dataset prioritizes real-world challenges,
including various environments (e.g., wild, indoor, urban, aerial view), intense camera motions (e.g., motion blur, frequent rotations, complex trajectories), and dynamic lighting conditions (e.g., underexposure or overexposure). 
The collected sequences are sourced from three streams: (1) 3 sequences from 360VOTS~\cite{360VOTS} meeting strict VO suitability criteria (stable ego-motion, appropriate motion blur), (2) 15 internet-sourced videos covering extreme scenarios (e.g., snow, nighttime, crowded markets), and (3) 2 long sequences recorded on a Hong Kong tramway. All raw videos are processed at 10 FPS and standardized to 3840$\times$1920 resolution, while
the sequences of sufficient length (i.e., $\geq 500$ frames) and continuous ego-motion are chosen with higher priorities for better VO tracking applicability.  
The collected in-the-wild videos lack true motion trajectories; therefore, we reconstruct pseudo ground truths via SfM software, such as COLMAP~\cite{colmap1, colmap2} and Agisoft Metashape~\cite{metashape}.
We assess the accuracy of the pseudo ground truth by comparing COLMAP and Metashape on TartanAirV2~\cite{tartanairV2}. 
The results show that Agisoft Metashape consistently produces trajectories with negligible ATE-RMSE (i.e., 0.027), which closely matches the provided ground truth, whereas COLMAP yields larger errors and drift (i.e., 2.535). Therefore, we adopt Agisoft Metashape as the pseudo ground-truth generation medium for our dataset.

Our final 360DVO dataset comprises 20 sequences with an average length of approximately 1k frames, which are partitioned into \textit{Easy} and \textit{Hard} subsets (10 each) based on trajectory complexity and environmental dynamics. \textit{Easy} sequences feature linear motions in static scenes (e.g., straight streets, open roads), while \textit{Hard} sequences incorporate aggressive rotations, rapid illumination changes (e.g., entering tunnels), and dynamic occlusions (e.g., crowds, vegetation). Representative frames are shown in Fig.~\ref{fig:dataset}. The dataset’s multi-domain coverage and explicit challenge stratification are designed to rigorously evaluate VO systems’ robustness against real-world perturbations, complementing existing synthetic benchmarks.

\vspace{-0.2cm}
\section{Experiments}\label{sec:experiments}
\subsection{Implementation Details}\label{sec:implementation}
360DVO is implemented in PyTorch with CUDA. With 62 scenes from TartanAirV2~\cite{tartanairV2} for training, the remaining 4 scenes are held out for evaluation. From each training scene, one easy and two hard sequences are selected, resulting in a training set of over 220k equirectangular images with depth maps. The system is trained end‑to‑end, supervised by relative pose loss and optical flow loss computed from ground‑truth depth. Following the training setup of DPVO~\cite{DPVO}, training runs for 100k iterations on a single RTX‑4090D GPU and takes approximately two days.

{
\setlength{\tabcolsep}{2pt}
\begin{table*}[t]
    \centering
    \caption{\changed{Comparison of ATE and RPE (translation/rotation) on TartanAirV2 evaluation set. Best result per sequence is marked in \textbf{\dovored{red}}.}}
\resizebox{\linewidth}{!}{
    \begin{tabular}{l cc cc cc cc cc}
    \toprule
    \multirow{2}{*}{Methods} & \multicolumn{2}{c}{CountryHouse} & \multicolumn{2}{c}{House} & \multicolumn{2}{c}{OldTownNight} & \multicolumn{2}{c}{VictorianStreet} & \multirow{2}{*}{Avg.} & \multirow{2}{*}{Succ.} \\ 
    \cmidrule(lr){2-3} \cmidrule(lr){4-5} \cmidrule(lr){6-7} \cmidrule(lr){8-9}
    & Easy & Hard & Easy & Hard & Easy & Hard & Easy & Hard && \\
    \midrule 
    ORB-SLAM3~\cite{ORBSLAM3}&\colorbox{white}{1.759}/\colorbox{white}{0.269}/\colorbox{white}{5.336}&\colorbox{white}{1.421}/\colorbox{white}{0.311}/\colorbox{white}{10.96}&\colorbox{white}{3.901}/\colorbox{white}{0.251}/\colorbox{white}{6.744}&\tableholder&\colorbox{white}{7.828}/\colorbox{white}{0.470}/\colorbox{white}{5.613}&\tableholder&\colorbox{white}{1.317}/\colorbox{white}{0.351}/\colorbox{white}{2.995}&\colorbox{white}{7.745}/\colorbox{white}{0.692}/\colorbox{white}{12.62}&\tableholder& 65\%\\
    Droid-SLAM~\cite{DroidSLAM}& \colorbox{white}{0.006}/\colorbox{white}{0.096}/\colorbox{white}{5.000}&\colorbox{white}{0.014}/\colorbox{white}{0.107}/\colorbox{white}{5.978}&\colorbox{firstcolor}{0.007}/\colorbox{white}{0.047}/\colorbox{white}{2.846}&\colorbox{white}{1.392}/\colorbox{white}{0.302}/\colorbox{white}{4.186}&\colorbox{white}{2.792}/\colorbox{white}{0.102}/\colorbox{white}{2.754}&\colorbox{white}{11.18}/\colorbox{white}{0.365}/\colorbox{white}{12.34}&\colorbox{white}{0.019}/\colorbox{white}{0.049}/\colorbox{white}{2.134}&\colorbox{white}{0.202}/\colorbox{white}{0.345}/\colorbox{white}{5.985}&\colorbox{white}{1.951}/\colorbox{white}{0.177}/\colorbox{white}{5.152}& 100\%\\
    DPVO~\cite{DPVO}&\colorbox{white}{0.027}/\colorbox{white}{0.097}/\colorbox{white}{5.003}&\colorbox{white}{0.359}/\colorbox{white}{0.102}/\colorbox{white}{5.693}&\colorbox{white}{0.604}/\colorbox{white}{0.062}/\colorbox{white}{2.858}&\colorbox{white}{2.749}/\colorbox{white}{0.208}/\colorbox{white}{4.378}&\colorbox{white}{0.196}/\colorbox{white}{0.107}/\colorbox{white}{2.940}&\colorbox{white}{8.138}/\colorbox{white}{0.347}/\colorbox{white}{12.17}&\colorbox{white}{0.049}/\colorbox{white}{0.051}/\colorbox{white}{2.147}&\colorbox{white}{1.219}/\colorbox{white}{0.241}/\colorbox{white}{5.683}&\colorbox{white}{1.668}/\colorbox{white}{0.152}/\colorbox{white}{5.110}& 100\% \\
    DPV-SLAM~\cite{dpvslam}&\colorbox{white}{0.008}/\colorbox{white}{0.097}/\colorbox{white}{5.001}&\colorbox{white}{0.149}/\colorbox{white}{0.193}/\colorbox{white}{12.30}&\colorbox{white}{0.409}/\colorbox{white}{0.053}/\colorbox{white}{2.581}&\colorbox{white}{0.056}/\colorbox{white}{0.191}/\colorbox{white}{12.80}&\colorbox{white}{0.493}/\colorbox{white}{0.106}/\colorbox{white}{2.915}&\colorbox{firstcolor}{0.104}/\colorbox{white}{0.191}/\colorbox{white}{4.905}&\colorbox{white}{0.063}/\colorbox{white}{0.050}/\colorbox{white}{2.143}&\colorbox{white}{0.065}/\colorbox{white}{0.178}/\colorbox{white}{8.706}&\colorbox{white}{0.168}/\colorbox{white}{0.132}/\colorbox{white}{6.420}& 100\% \\
    \midrule 
    OpenVSLAM~\cite{openvslam}& \colorbox{white}{0.283}/\colorbox{white}{0.007}/\colorbox{white}{0.376}&\tableholder&\colorbox{white}{0.020}/\colorbox{white}{0.007}/\colorbox{white}{0.090}&\tableholder&\colorbox{firstcolor}{0.033}/\colorbox{white}{0.019}/\colorbox{white}{0.355}&\tableholder&\colorbox{firstcolor}{0.016}/\colorbox{white}{0.010}/\colorbox{white}{0.060}&\tableholder&\tableholder&53\% \\
    360VO~\cite{360VO}& \colorbox{white}{2.277}/\colorbox{white}{0.184}/\colorbox{white}{3.869}&\colorbox{white}{1.977}/\colorbox{white}{0.238}/\colorbox{white}{9.066}&\tableholder&\tableholder& \tableholder&\tableholder&\tableholder&\tableholder&\tableholder&27\% \\
    360DVO (ours)& \colorbox{firstcolor}{0.004}/\colorbox{firstcolor}{0.001}/\colorbox{firstcolor}{0.010}&\colorbox{firstcolor}{0.005}/\colorbox{firstcolor}{0.001}/\colorbox{firstcolor}{0.020}&\colorbox{white}{0.025}/\colorbox{firstcolor}{0.002}/\colorbox{firstcolor}{0.016}&\colorbox{firstcolor}{0.038}/\colorbox{firstcolor}{0.005}/\colorbox{firstcolor}{0.037}&\colorbox{white}{0.039}/\colorbox{firstcolor}{0.003}/\colorbox{firstcolor}{0.009}&\colorbox{white}{0.145}/\colorbox{firstcolor}{0.031}/\colorbox{firstcolor}{0.038}&\colorbox{white}{0.025}/\colorbox{firstcolor}{0.002}/\colorbox{firstcolor}{0.010}&\colorbox{firstcolor}{0.020}/\colorbox{firstcolor}{0.002}/\colorbox{firstcolor}{0.013}&\colorbox{firstcolor}{0.038}/\colorbox{firstcolor}{0.006}/\colorbox{firstcolor}{0.019} &100\% \\
    \bottomrule
    \end{tabular}
 }
    \label{tab:exp_tartan}
\end{table*}
}

\vspace{-0.2cm}
\subsection{Evaluation}\label{sec:evaluation}
We compare 360DVO with state‑of‑the‑art pinhole methods and OVO baselines, OpenVSLAM~\cite{openvslam} and 360VO~\cite{360VO}, on 360VO dataset~\cite{360VO}, TartanAirV2~\cite{tartanairV2} evaluation set, and our new proposed 360DVO dataset. Following Sim(3) Umeyama alignment~\cite{umeyama} of the predicted trajectories to the ground truth, the root mean squared error (RMSE) is reported for three metrics: absolute trajectory error (ATE, in meters), translational relative pose error (RPE(t), in m/frame), and rotational relative pose error (RPE(r), in deg/frame). Results are averaged over five runs. For timestamp mismatches, evaluation is restricted to overlapping trajectory segments. In each run, if tracking is lost in more than half of the frames in a sequence, the run is counted as a failure for that sequence. Cases that fail in more than half of the five runs are indicated as $``-"$.

\noindent{\textbf{360VO Dataset.}} The 360VO dataset~\cite{360VO} is a relatively easy synthetic dataset with smooth camera motion and static illumination. 
As shown in Tab.~\ref{tab:exp_360vo}, 360DVO achieves better performance on all metrics than both OpenVSLAM and 360VO, reducing the ATE by 10\% compared with 360VO.

\noindent{\textbf{TartanAirV2 Evaluation Set.}}
As noted above, we construct an evaluation set from 4 scenes in TartanAirV2~\cite{tartanairV2}, spanning day/night and indoor/outdoor settings. We select one easy and one hard sequence per scene, with hard sequences exhibiting stronger rotations. We also evaluate pinhole baselines on the corresponding perspective images provided by TartanAirV2.
As Tab.~\ref{tab:exp_tartan} shows, our 360DVO with default settings attains the highest success rate and accuracy, whereas OpenVSLAM and 360VO nearly fail on all hard sequences. 
Learning-based pinhole methods~\cite{DroidSLAM,DPVO,dpvslam}, trained on TartanAir~\cite{tartanair}, perform reasonably well but struggle on the Hard sequences, highlighting the benefit of omnidirectional FOVs for visual odometry.
These results highlight 360DVO’s robustness to challenging scenarios involving intense camera rotations.

\begin{table}[t]
    \centering
    \caption{Comparison of average ATE and RPE (translation/rotation) on 360VO synthetic dataset.}
    \begin{tabular}{l|c c c}
    \toprule
    Methods & ATE & RPE(t) & RPE(r) \\
    \midrule 
    OpenVSLAM~\cite{openvslam}&2.25&0.312&0.452 \\
    360VO~\cite{360VO}&1.24&0.291&0.455  \\
    360DVO (ours)&\colorbox{firstcolor}{1.11}&\colorbox{firstcolor}{0.235}&\colorbox{firstcolor}{0.440} \\
    \bottomrule
    \end{tabular}
    \label{tab:exp_360vo}
    \vspace{-0.2cm}
\end{table}

{
\setlength{\tabcolsep}{0pt}
\begin{table*}[t!]
    \centering
    \caption{
    Quantitative comparison of trajectory accuracy (ATE/RPE(t)/RPE(r)) and tracking success rate (\%) on the 360DVO dataset (\textit{Easy} and \textit{Hard}). $``-"$ indicates failure. Best result per sequence is in \textbf{\dovored{red}}, second-best result in \textbf{\openvblue{blue}}. 360DVO (fast) denotes the modification of using lower resolution images as input while maintaining sparser sampling patches. All results are shown with 2-decimal precision for clarity.
    }
    \tabcolsep=6pt
\resizebox{\linewidth}{!}{ 
    \begin{tabular}{l ccccc ccccc c c}
    \toprule
    \multirow{2}{*}{Methods} & \multicolumn{10}{c}{Easy Sequences} & \multirow{2}{*}{Avg.} & \multirow{2}{*}{Succ.}\\ \cmidrule(lr){2-11}
        & 00&01&02&03&04&05&06&07&08&09&  \\
    \midrule 
        ORB-SLAM3~\cite{ORBSLAM3}& \colorbox{white}{24.2}/\colorbox{white}{0.86}/\colorbox{white}{4.43}&\colorbox{white}{0.33}/\colorbox{white}{0.44}/\colorbox{white}{0.11}&\tableholdern&\tableholdern&\colorbox{white}{3.75}/\colorbox{white}{0.23}/\colorbox{white}{0.11}& \tableholdern&\colorbox{white}{0.47}/\colorbox{white}{0.29}/\colorbox{white}{0.53}&\tableholdern&\tableholdern&\tableholdern&\tableholdern&46\% \\ 
        Droid-SLAM~\cite{DroidSLAM}& \colorbox{white}{2.28}/\colorbox{white}{0.40}/\colorbox{white}{0.35}&\colorbox{white}{0.61}/\colorbox{firstcolor}{0.13}/\colorbox{firstcolor}{0.04}&\colorbox{white}{44.2}/\colorbox{firstcolor}{1.33}/\colorbox{white}{0.19}&\colorbox{white}{22.7}/\colorbox{firstcolor}{1.08}/\colorbox{white}{0.69}&\colorbox{white}{0.57}/\colorbox{firstcolor}{0.02}/\colorbox{secondcolor}{0.03}& \colorbox{white}{5.27}/\colorbox{firstcolor}{0.15}/\colorbox{white}{0.07}&\colorbox{white}{0.37}/\colorbox{white}{0.22}/\colorbox{firstcolor}{0.41}&\tableholdern&23.7/\colorbox{firstcolor}{0.72}/\colorbox{white}{1.58}&\colorbox{white}{2.70}/\colorbox{firstcolor}{0.04}/\colorbox{white}{0.09}&\tableholdern&90\%  \\ 
        DPVO~\cite{DPVO}& \colorbox{white}{1.37}/\colorbox{white}{0.32}/\colorbox{white}{0.11}&\colorbox{firstcolor}{0.13}/\colorbox{white}{0.44}/\colorbox{white}{0.11}&\colorbox{white}{37.5}/\colorbox{secondcolor}{1.36}/\colorbox{white}{0.13}&\colorbox{white}{6.91}/\colorbox{white}{1.33}/\colorbox{white}{0.46}&\colorbox{white}{0.57}/\colorbox{white}{0.23}/\colorbox{white}{0.03}& \colorbox{white}{4.89}/\colorbox{secondcolor}{0.24}/\colorbox{white}{0.07}&\colorbox{white}{0.35}/\colorbox{white}{0.21}/\colorbox{white}{0.52}&\colorbox{white}{16.4}/\colorbox{white}{1.68}/\colorbox{white}{0.15}&\colorbox{white}{7.93}/\colorbox{white}{0.88}/\colorbox{white}{0.44}&\colorbox{white}{2.23}/\colorbox{white}{0.41}/\colorbox{white}{0.10}&\colorbox{white}{7.83}/\colorbox{white}{0.71}/\colorbox{white}{0.21}&100\%  \\ 
        DPV-SLAM~\cite{dpvslam}&\colorbox{secondcolor}{1.19}/\colorbox{secondcolor}{0.32}/\colorbox{white}{0.11}&\colorbox{white}{0.23}/\colorbox{white}{0.44}/\colorbox{white}{0.10}&\colorbox{white}{33.7}/\colorbox{white}{1.38}/\colorbox{white}{0.13}&\colorbox{white}{6.74}/\colorbox{white}{1.34}/\colorbox{white}{0.46}&\colorbox{secondcolor}{0.50}/\colorbox{white}{0.23}/\colorbox{firstcolor}{0.03}&\colorbox{white}{4.97}/\colorbox{white}{0.25}/\colorbox{white}{0.07}&\colorbox{secondcolor}{0.30}/\colorbox{white}{0.21}/\colorbox{white}{0.52}&\colorbox{white}{18.3}/\colorbox{white}{1.68}/\colorbox{white}{0.13}&\colorbox{white}{7.31}/\colorbox{white}{1.02}/\colorbox{white}{0.60}&\colorbox{white}{2.36}/\colorbox{white}{0.40}/\colorbox{white}{0.10}&\colorbox{white}{7.56}/\colorbox{white}{0.73}/\colorbox{white}{0.22}&100\% \\
    \midrule
        OpenVSLAM~\cite{openvslam}& \colorbox{white}{1.73}/\colorbox{white}{0.33}/\colorbox{white}{0.11}&\colorbox{white}{0.38}/\colorbox{white}{0.44}/\colorbox{white}{0.08}&\colorbox{firstcolor}{18.5}/\colorbox{white}{1.36}/\colorbox{white}{0.11}&\colorbox{firstcolor}{2.77}/\colorbox{white}{1.44}/\colorbox{white}{0.45}&\colorbox{white}{4.36}/\colorbox{white}{0.23}/\colorbox{white}{0.04}& \colorbox{white}{0.71}/\colorbox{white}{0.25}/\colorbox{secondcolor}{0.03}&\colorbox{white}{0.49}/\colorbox{white}{0.21}/\colorbox{secondcolor}{0.49}&\colorbox{secondcolor}{11.9}/\colorbox{white}{1.68}/\colorbox{secondcolor}{0.13}&\colorbox{secondcolor}{3.11}/\colorbox{white}{0.89}/\colorbox{white}{0.27}&\colorbox{firstcolor}{1.66}/\colorbox{white}{0.41}/\colorbox{white}{0.11}&\colorbox{secondcolor}{4.57}/\colorbox{white}{0.72}/\colorbox{white}{0.18}&94\%  \\
        360VO~\cite{360VO}& \colorbox{white}{15.1}/\colorbox{white}{0.37}/\colorbox{white}{0.38}&\colorbox{white}{14.3}/\colorbox{white}{1.00}/\colorbox{white}{1.02}&\tableholdern&\tableholdern&\tableholdern& \tableholdern&\colorbox{white}{15.0}/\colorbox{white}{0.28}/\colorbox{white}{0.98}&\colorbox{white}{91.8}/\colorbox{white}{3.28}/\colorbox{white}{2.12}&\colorbox{white}{84.1}/\colorbox{white}{1.84}/\colorbox{white}{2.41}&\colorbox{white}{24.1}/\colorbox{white}{0.53}/\colorbox{white}{1.17}&\tableholdern&42\% \\ 
        360DVO (default)& \colorbox{firstcolor}{1.06}/\colorbox{firstcolor}{0.32}/\colorbox{firstcolor}{0.09}&\colorbox{secondcolor}{0.22}/\colorbox{secondcolor}{0.43}/\colorbox{secondcolor}{0.07}&\colorbox{secondcolor}{22.5}/\colorbox{white}{1.36}/\colorbox{firstcolor}{0.07}&\colorbox{secondcolor}{3.26}/\colorbox{secondcolor}{1.32}/\colorbox{firstcolor}{0.44}&\colorbox{firstcolor}{0.45}/\colorbox{secondcolor}{0.22}/\colorbox{white}{0.05}& \colorbox{firstcolor}{0.50}/\colorbox{white}{0.24}/\colorbox{firstcolor}{0.02}&\colorbox{firstcolor}{0.28}/\colorbox{firstcolor}{0.20}/\colorbox{white}{0.51}&\colorbox{firstcolor}{1.08}/\colorbox{secondcolor}{1.67}/\colorbox{firstcolor}{0.10}&\colorbox{firstcolor}{1.82}/\colorbox{secondcolor}{0.86}/\colorbox{firstcolor}{0.06}&\colorbox{secondcolor}{2.00}/\colorbox{secondcolor}{0.40}/\colorbox{secondcolor}{0.09}&\colorbox{firstcolor}{3.31}/\colorbox{firstcolor}{0.70}/\colorbox{firstcolor}{0.15} &100\%\\
        360DVO (fast)& \colorbox{white}{1.90}/\colorbox{white}{0.32}/\colorbox{secondcolor}{0.09}&\colorbox{white}{0.66}/\colorbox{white}{0.44}/\colorbox{white}{0.09}&\colorbox{white}{23.4}/\colorbox{white}{1.37}/\colorbox{secondcolor}{0.07}&\colorbox{white}{3.53}/\colorbox{white}{1.33}/\colorbox{secondcolor}{0.45}&\colorbox{white}{5.02}/\colorbox{white}{0.23}/\colorbox{white}{0.12}&\colorbox{secondcolor}{0.62}/\colorbox{white}{0.24}/\colorbox{white}{0.05}&\colorbox{white}{0.45}/\colorbox{secondcolor}{0.21}/\colorbox{white}{0.51}&\colorbox{white}{19.6}/\colorbox{firstcolor}{1.67}/\colorbox{white}{0.19}&\colorbox{white}{4.25}/\colorbox{white}{0.86}/\colorbox{secondcolor}{0.07}&\colorbox{white}{2.52}/\colorbox{white}{0.40}/\colorbox{firstcolor}{0.09}&\colorbox{white}{6.20}/\colorbox{secondcolor}{0.71}/\colorbox{secondcolor}{0.17}&100\%\\
    \midrule \midrule
    \multirow{2}{*}{Methods} & \multicolumn{10}{c}{Hard Sequences} & \multirow{2}{*}{Avg.}\\ \cmidrule(lr){2-11}
        & 00&01&02&03&04&05&06&07&08&09& \\
    \midrule
        ORB-SLAM3~\cite{ORBSLAM3}& \colorbox{white}{0.40}/\colorbox{white}{0.71}/\colorbox{white}{0.13}&\colorbox{white}{12.9}/\colorbox{white}{0.56}/\colorbox{white}{6.05}&\colorbox{white}{3.49}/\colorbox{white}{0.94}/\colorbox{white}{0.73}&\colorbox{white}{7.84}/\colorbox{white}{0.32}/\colorbox{white}{9.83}&\colorbox{white}{6.84}/\colorbox{white}{0.79}/\colorbox{white}{3.69}&\colorbox{white}{3.11}/ \colorbox{white}{1.18}/\colorbox{white}{6.50}&\tableholdern&\tableholdern&\tableholdern&\colorbox{white}{37.0}/\colorbox{white}{6.60}/\colorbox{firstcolor}{1.53}&\tableholdern&68\% \\
        Droid-SLAM~\cite{DroidSLAM}& \colorbox{secondcolor}{0.20}/\colorbox{firstcolor}{0.02}/\colorbox{white}{0.04}&\colorbox{white}{2.93}/\colorbox{white}{0.60}/\colorbox{white}{2.81}&\colorbox{white}{2.27}/\colorbox{firstcolor}{0.20}/\colorbox{white}{0.65}&\colorbox{white}{2.42}/\colorbox{white}{0.33}/\colorbox{white}{0.43}&\colorbox{white}{10.2}/\colorbox{white}{0.77}/\colorbox{white}{0.60}& \colorbox{secondcolor}{2.72}/\colorbox{white}{0.88}/\colorbox{white}{1.41}&\colorbox{firstcolor}{12.9}/\colorbox{white}{0.61}/\colorbox{white}{0.64}&\colorbox{white}{3.50}/\colorbox{firstcolor}{0.94}/\colorbox{secondcolor}{0.66}&\colorbox{white}{5.48}/\colorbox{firstcolor}{1.25}/\colorbox{white}{0.13}&\colorbox{white}{29.4}/\colorbox{white}{1.27}/\colorbox{white}{1.54}&\colorbox{white}{7.20}/\colorbox{firstcolor}{0.69}/\colorbox{white}{0.89}&100\% \\ 
        DPVO~\cite{DPVO}& \colorbox{white}{0.72}/\colorbox{white}{0.71}/\colorbox{white}{0.03}&\colorbox{white}{10.4}/\colorbox{white}{0.37}/\colorbox{white}{1.96}&\colorbox{white}{1.34}/\colorbox{white}{0.91}/\colorbox{white}{0.61}&\colorbox{white}{0.71}/\colorbox{white}{0.38}/\colorbox{secondcolor}{0.10}&\colorbox{white}{0.61}/\colorbox{white}{0.59}/\colorbox{white}{0.06}&\colorbox{white}{4.03}/ \colorbox{white}{1.03}/\colorbox{white}{1.33}&\colorbox{white}{19.3}/\colorbox{white}{0.53}/\colorbox{white}{0.26}&\colorbox{secondcolor}{1.66}/\colorbox{white}{1.19}/\colorbox{firstcolor}{0.01}&\colorbox{white}{7.97}/\colorbox{white}{1.69}/\colorbox{white}{0.12}&\colorbox{white}{22.5}/\colorbox{white}{1.06}/\colorbox{white}{1.63}&\colorbox{white}{6.92}/\colorbox{white}{0.85}/\colorbox{secondcolor}{0.71}&100\% \\
        DPV-SLAM~\cite{dpvslam}&\colorbox{white}{0.64}/\colorbox{white}{0.71}/\colorbox{white}{0.03}&\colorbox{white}{9.14}/\colorbox{white}{0.32}/\colorbox{white}{2.51}&\colorbox{white}{0.95}/\colorbox{white}{0.99}/\colorbox{white}{0.61}&\colorbox{secondcolor}{0.70}/\colorbox{white}{0.39}/\colorbox{white}{0.14}&\colorbox{white}{0.58}/\colorbox{white}{0.60}/\colorbox{white}{0.06}&\colorbox{white}{3.54}/\colorbox{white}{1.04}/\colorbox{white}{1.32}&\colorbox{secondcolor}{17.7}/\colorbox{white}{0.54}/\colorbox{white}{0.22}&\colorbox{white}{2.09}/\colorbox{white}{1.20}/\colorbox{white}{1.02}&\colorbox{white}{6.80}/\colorbox{secondcolor}{1.68}/\colorbox{white}{0.12}&\colorbox{white}{28.4}/\colorbox{firstcolor}{0.98}/\colorbox{white}{1.74}&\colorbox{white}{7.05}/\colorbox{white}{0.84}/\colorbox{white}{0.78}&100\%\\
    \midrule
        OpenVSLAM~\cite{openvslam}& \colorbox{firstcolor}{0.20}/\colorbox{white}{0.71}/\colorbox{white}{0.04}&\colorbox{secondcolor}{1.68}/\colorbox{white}{0.27}/\colorbox{firstcolor}{0.24}&\colorbox{white}{4.41}/\colorbox{secondcolor}{0.87}/\colorbox{white}{0.65}&\colorbox{firstcolor}{0.11}/\colorbox{white}{0.39}/\colorbox{firstcolor}{0.07}&\colorbox{secondcolor}{0.28}/\colorbox{white}{0.60}/\colorbox{secondcolor}{0.06}& \colorbox{white}{10.3}/\colorbox{white}{0.98}/\colorbox{white}{5.35}&\colorbox{white}{21.4}/\colorbox{white}{0.64}/\colorbox{white}{0.18}&\colorbox{white}{18.6}/\colorbox{white}{1.23}/\colorbox{white}{1.00}&\colorbox{white}{5.17}/\colorbox{white}{1.74}/\colorbox{white}{0.16}&\colorbox{white}{14.8}/\colorbox{white}{1.14}/\colorbox{secondcolor}{1.54}&\colorbox{white}{7.69}/\colorbox{white}{0.86}/\colorbox{white}{0.93}&92\%  \\
        360VO~\cite{360VO}& \tableholdern&\tableholdern&\colorbox{white}{19.1}/\colorbox{white}{0.93}/\colorbox{white}{1.13}&\tableholdern&\colorbox{white}{3.43}/\colorbox{firstcolor}{0.38}/\colorbox{white}{0.50}& \colorbox{white}{10.3}/\colorbox{firstcolor}{0.57}/\colorbox{firstcolor}{0.17}&\tableholdern&\colorbox{white}{34.3}/\colorbox{white}{1.73}/\colorbox{white}{1.65}&\colorbox{white}{105}/\colorbox{white}{3.73}/\colorbox{white}{1.30}&\tableholdern&\tableholdern&44\% \\ 
        360DVO (default)& \colorbox{white}{0.45}/\colorbox{white}{0.70}/\colorbox{firstcolor}{0.02}&\colorbox{white}{4.97}/\colorbox{secondcolor}{0.23}/\colorbox{white}{1.01}&\colorbox{firstcolor}{0.61}/\colorbox{white}{0.90}/\colorbox{firstcolor}{0.60}&\colorbox{white}{1.18}/\colorbox{secondcolor}{0.31}/\colorbox{white}{0.31}&\colorbox{white}{5.10}/\colorbox{secondcolor}{0.41}/\colorbox{white}{0.32}&\colorbox{white}{2.82}/\colorbox{secondcolor}{0.86}/\colorbox{white}{4.39}&\colorbox{white}{23.1}/\colorbox{firstcolor}{0.49}/\colorbox{firstcolor}{0.15}&\colorbox{firstcolor}{1.50}/\colorbox{white}{1.18}/\colorbox{white}{1.06}&\colorbox{secondcolor}{3.33}/\colorbox{white}{1.69}/\colorbox{firstcolor}{0.11}&\colorbox{firstcolor}{0.36}/\colorbox{secondcolor}{1.03}/\colorbox{white}{1.55}&\colorbox{secondcolor}{4.35}/\colorbox{secondcolor}{0.78}/\colorbox{white}{0.95}&100\% \\
        360DVO (fast)& \colorbox{white}{0.82}/\colorbox{secondcolor}{0.70}/\colorbox{secondcolor}{0.02}&\colorbox{firstcolor}{0.81}/\colorbox{firstcolor}{0.22}/\colorbox{secondcolor}{0.87}&\colorbox{secondcolor}{0.61}/\colorbox{white}{0.90}/\colorbox{secondcolor}{0.60}&\colorbox{white}{2.67}/\colorbox{firstcolor}{0.31}/\colorbox{white}{0.15}&\colorbox{firstcolor}{0.19}/\colorbox{white}{0.42}/\colorbox{firstcolor}{0.04}&\colorbox{firstcolor}{2.71}/\colorbox{white}{1.05}/\colorbox{secondcolor}{1.30}&\colorbox{white}{22.7}/\colorbox{secondcolor}{0.49}/\colorbox{secondcolor}{0.16}&\colorbox{white}{2.46}/\colorbox{secondcolor}{1.17}/\colorbox{white}{1.08}&\colorbox{firstcolor}{3.08}/\colorbox{white}{1.69}/\colorbox{secondcolor}{0.11}&\colorbox{secondcolor}{0.69}/\colorbox{white}{1.03}/\colorbox{white}{1.55}&\colorbox{firstcolor}{3.68}/\colorbox{white}{0.80}/\colorbox{firstcolor}{0.59}&100\% \\
    \bottomrule
    \end{tabular}
}
    \label{tab:exp_360vod}
\end{table*}
}

\begin{figure*}
    \centering
    \includegraphics[width=0.99\linewidth]{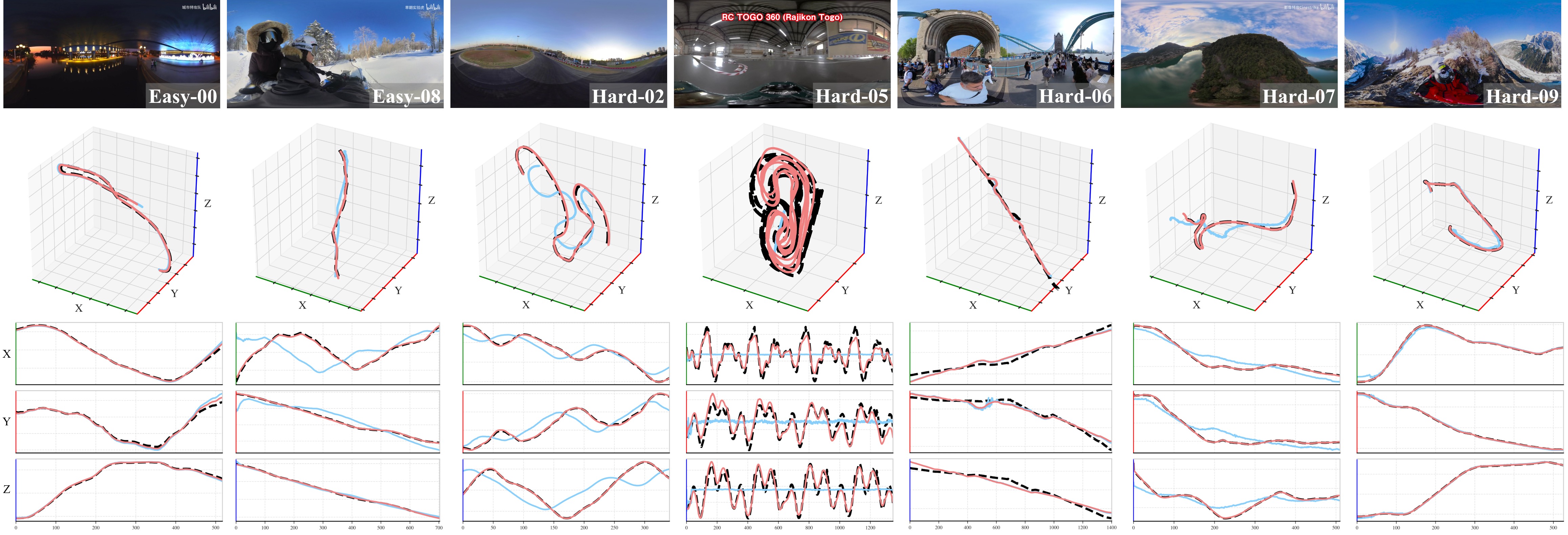}
    \caption{Trajectories Comparison on the 360DVO dataset in 3D space, with position variations along the X, Y, and Z axes plotted over all frames. The ground truth is shown in \textbf{black dashed lines}, 360DVO results in \textbf{\dovored{red solid lines}}, and OpenVSLAM results in \textbf{\openvblue{blue solid lines}}.
    }
    \vspace{-0.3cm}
    \label{fig:res}
\end{figure*}

\noindent{\textbf{360DVO Dataset. }} 
To decouple the benefits of a wide field of view (FOV) from algorithmic performance, we evaluate existing pinhole methods alongside OVO methods on our 360DVO dataset. 
Virtual monocular sequences are derived by extracting 90-degree FOV perspective images of size $640\times 640$ from the omnidirectional images. SOTA pinhole methods are benchmarked on this branch, including ORB-SLAM3~\cite{ORBSLAM3}, Droid-SLAM~\cite{DroidSLAM}, DPVO~\cite{DPVO}, and DPV-SLAM~\cite{dpvslam}. For the pinhole methods, the camera intrinsics are uniformly set to the ideal values (e.g., $f_x=f_y=c_x=c_y=320$). For OVO methods constrained by the spherical camera model, the intrinsics are computed from the input resolution as in Eq.~\ref{eq:intrinsic}.

Narrow FOVs make challenging scenes harder by increasing feature leave-view events and shortening long-term co-visibility.
Overall, OVO methods demonstrate more stable and higher performance than pinhole methods shown in Tab.~\ref{tab:exp_360vod}, revealing the inherent value of 360-degree coverage. 
360DVO reduces ATE over DPV-SLAM by 56.2\% on \textit{Easy} and outperforms DPVO with 37.1\% improvement on \textit{Hard}.
Notably, the pinhole methods outperform the OVO methods on Hard‑06 captured on a crowded bridge. The pinhole crops predominantly contain static bridge structures, whereas the omnidirectional images include many dynamic objects, which degrades feature matching performance.

Additionally, learning-based methods outperform classical ones on average. 
Among pinhole methods, learning-based DPV-SLAM is best on \textit{Easy} and DPVO outperforms others on \textit{Hard}, while Droid-SLAM is competitive but fails on Easy-07. 
On the omnidirectional side, 360DVO surpasses OpenVSLAM by 27.6\% on \textit{Easy} and 43.4\% on \textit{Hard}. OpenVSLAM is competitive on \textit{Easy} (i.e., 4.57) but underperforms on \textit{Hard} (i.e., 7.69). Direct method 360VO fails on many sequences, with an average success rate of 43\%, far below 360DVO's 100\%. Qualitatively, our 360DVO system demonstrates the capabilities for rapid 6-DoF motions, large rotations, and dynamic illumination from aerial views (e.g., Hard-02, Hard-07, Hard-09) compared with OpenVSLAM, as shown in Fig.~\ref{fig:res}. In the indoor cases (e.g., Hard-05), 360DVO remains competitive under the challenges of substantial motion blur.
These advancements highlight the advantages of the DAS-Feat module in extracting deep features from 360-degree context, and the stability provided by ODBA in optimization.

To further evaluate the efficiency of our proposed 360DVO, we establish two baselines with default and fast settings. 360DVO (default) executes under the configuration of inputs in $3840\times 1920$ size, 192 patches per image (i.e., $N$ in Eq.~\ref{eq:patchify}), and a gradient-based patch selection strategy. 360DVO (fast) is deployed by downsampling inputs to $1920\times 960$ and halving $N$ to 96, meeting real-time constraints. Detailed in Tab.~\ref{tab:exp_360vod}, 360DVO (default) achieves the best overall performance on the 360DVO dataset, ranking first on \textit{Easy} (i.e., 3.31) and second on \textit{Hard} (i.e., 4.35), while 360DVO (fast) performs best on \textit{Hard} (i.e., 3.68).
We also observe a characteristic failure mode for 360DVO (default) with the high-resolution setting. On sequence Hard-04, 360DVO (default) underperforms both classical and learning-based methods (e.g., OpenVSLAM~\cite{openvslam}, DPVO~\cite{DPVO}), while 360DVO (fast) achieves the best result (i.e., 0.190). 
This sequence is dominated by repetitive textures (e.g., sky, grass, foliage). High-resolution gradient-based selection admits many ambiguous patches, yielding unstable correspondences in new keyframes. The fast configuration implicitly regularizes the problem by reducing per-frame patches and feature scales, thereby avoiding low-entropy, look‑alike regions.

\begin{figure*}[t]
    \centering
    \captionsetup[subfloat]{labelfont={rm,footnotesize},textfont=footnotesize}
    \subfloat[\changed{ResNet Encoder} (accuracy: 45\%)\label{base_tracking}]{\includegraphics[width=0.99\linewidth]{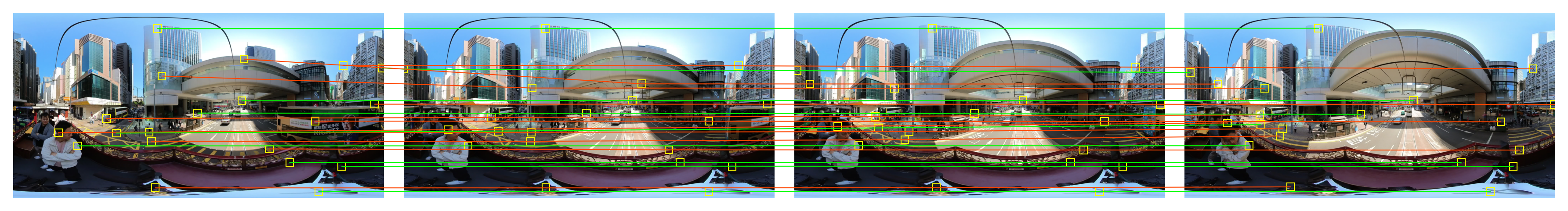}}\\
    \vspace{-0.3cm}\subfloat[SphereResNet (accuracy: 90\%)\label{sphericalresnet_tracking}]{\includegraphics[width=0.99\linewidth]{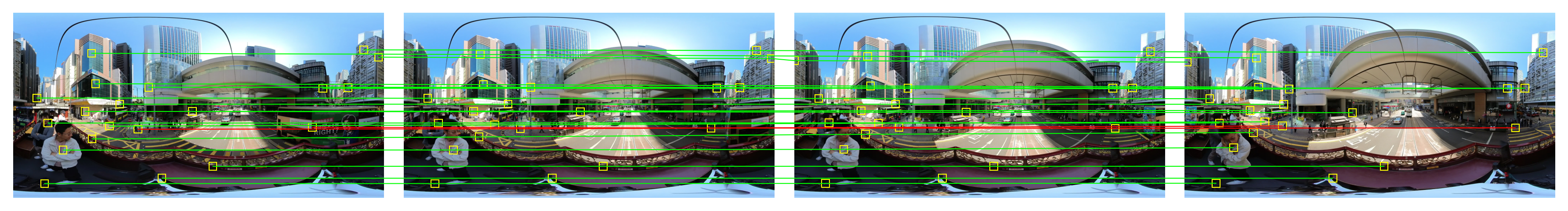}}
    \caption{
    Two samples of predicted patch trajectories (\changed{ResNet Encoder} vs. SphereResNet). Correct in green, incorrect in red. Patches are highlighted by yellow squares. The proposed SphereResNet yields higher tracking accuracy.}
    \vspace{-0.3cm}
    \label{fig:patch-tracking}
\end{figure*}

\vspace{-0.2cm}
\subsection{Ablation Studies}\label{sec:ablation}
We perform ablation studies of components and hyperparameters on our dataset, reported in Tab.~\ref{tab:abl} and Tab.~\ref{tab:abl_para}.  

\begin{table}[t]
    \centering
    \caption{Ablation study of our 360DVO system to verify the effectiveness of proposal components.
    }
\resizebox{1\linewidth}{!}{
    \begin{tabular}{c ccc ccc}
    \toprule
        \multirow{2}{*}{ID}&\multicolumn{3}{c}{Variants}&\multicolumn{3}{c}{ATE} \\
        \cmidrule(lr){2-4}\cmidrule(lr){5-7}
        &Input&Feature&BA&Easy&Hard&Avg. \\
    \midrule
        DPVO~\cite{DPVO}&Pinhole&\changed{ResNet\cite{resnet}}&DBA&7.83&6.92&7.37 \\
        \#1&$360\degree$&\changed{ResNet\cite{resnet}}&ODBA&7.33&12.7&9.99 \\
        \#2&$360\degree$&SphereNet~\cite{SphereNet}&ODBA&-&-&- \\
        360DVO&$360\degree$&SphereResNet&ODBA&3.31&4.35&3.83 \\
    \bottomrule
    \end{tabular}
}
    \vspace{-0.2cm}
    \label{tab:abl}
\end{table}

\noindent{\textbf{From Pinhole to Omnidirection. }}  
Replacing the BA module of DPVO~\cite{DPVO} with the novel omnidirectional differentiable bundle adjustment (ODBA) module enables 360-degree camera pose optimization. 
However, although the modification allows 360-degree image processing, the method relying on classic CNN features has degraded performance as reported in Tab.~\ref{tab:abl} (\#1). It highlights the necessity of a tailored network for the 360-degree image feature extraction. 

\noindent{\textbf{Distortion-Aware Spherical Features. }}
However, when replacing the classic CNN with a spherical feature extractor, SphereNet \cite{SphereNet}, the model suffers gradient explosions despite clipping gradient and tuning learning-rate during training. It fails to estimate any camera poses consequently, shown in Tab.~\ref{tab:abl} (\#2).
By contrast, the proposed SphereResNet is able to stably learn distortion‑resistant features and yield the best accuracy, proving its efficacy. The efficacy of SphereResNet has also been verified in Fig.~\ref{fig:patch-tracking}.

\begin{table}[t]
    \centering
    \caption{
    The influence of 360DVO hyperparameters in terms of trajectory accuracy.
    \#D denotes default setting, \#F denotes fast version. \#R, \#P, and \#S denote resolution, patch count ($N$), and patch selection, respectively.}
\resizebox{1\linewidth}{!}{
    \begin{tabular}{l ccc cccc}
    \toprule
        \multirow{2}{*}{ID}&\multicolumn{3}{c}{Settings}&\multicolumn{3}{c}{ATE}&\multirow{2}{*}{FPS} \\
        \cmidrule(lr){2-4}\cmidrule(lr){5-7}
        &Resolution&$N$&Selection&Easy&Hard&Avg.&\\
    \midrule \midrule
        \#D&3840$\times$1920&192&Gradient&3.31&4.35&3.83&8 \\
    \midrule
        \#R0&1920$\times$960&192&Gradient&5.58&3.75&4.66&17 \\
        \#R1&960$\times$480&192&Gradient&6.95&3.86&5.41&22 \\
    \midrule
        \#P0&3840$\times$1920&384&Gradient&3.86&6.82&5.34&5 \\
        \#P1&3840$\times$1920&96&Gradient&3.90&6.51&5.20&10\\
        \#P2&3840$\times$1920&48&Gradient&3.90&5.95&4.93&14 \\
    \midrule
        \#S&3840$\times$1920&192&Random&4.40&7.29&5.85&8 \\
    \midrule
        \#F&1920$\times$960&96&Gradient&6.20&3.68&4.94&27 \\
    \bottomrule
    \end{tabular}
}
\vspace{-0.3cm}
    \label{tab:abl_para}
\end{table}

\noindent{\textbf{Hyperparameters. }}  
Tab.~\ref{tab:abl_para} reveals clear trade-offs between accuracy and runtime efficiency by various hyperparameters. 
Downscaling the default resolution to $1920\times960$ ($\#$R0) and $960\times480$ ($\#$R1) raises FPS from 8 to 17 and 22, while worsening average ATE by 21.6\% and 41.0\%. 
Low-resolution settings reduce apparent pixel displacement and blur, stabilizing aggressive motion but sacrificing details in general scenarios. 
The default setting ($\#$D in Tab.~\ref{tab:abl_para}) with patch count $N=192$ reaches an optimal result. Increased to 384 per-frame patches ($\#$P0), the BA module is overconstrained to ambiguous correspondences. The computational cost increases while the estimated trajectory accuracy decreases. When cutting down $N$ to 96 ($\#$P1) and 48 ($\#$P2), the extracted features are underrepresented across the entire image, affecting accuracy as well. In addition, the gradient-based strategy outperforms the random-based $\#$S, concentrating patches on high-entropy regions.
The fast variant of $\#$F in Tab.~\ref{tab:abl_para} corresponding to 360DVO (fast) in Tab.~\ref{tab:exp_360vod}, achieves real-time performance at 27 FPS with a competitive accuracy (i.e., 4.94). 

\subsection{Edge Deployment and Limitation}\label{sec:edge}

\begin{table}[t]
    \centering
    \caption{\changed{Runtime (fps) and Performance (ATE, m) comparisons on Jetson Orin.}
    }
    \begin{tabular}{l|c c c c}
    \toprule
    Methods & Easy & Hard & Avg. & Fps \\
    \midrule 
    OpenVSLAM~\cite{openvslam}&6.01&6.75&6.38& {3-7} \\
    360DVO (default)&{3.46}&4.49&{3.98}&2 \\
    360DVO (fast)&6.19&{3.95}&5.07&5 \\
    \bottomrule
    \end{tabular}
    \vspace{-0.3cm}
    \label{tab:jetson}
\end{table}

\begin{figure}[t]
    \centering
    \includegraphics[width=0.99\linewidth]{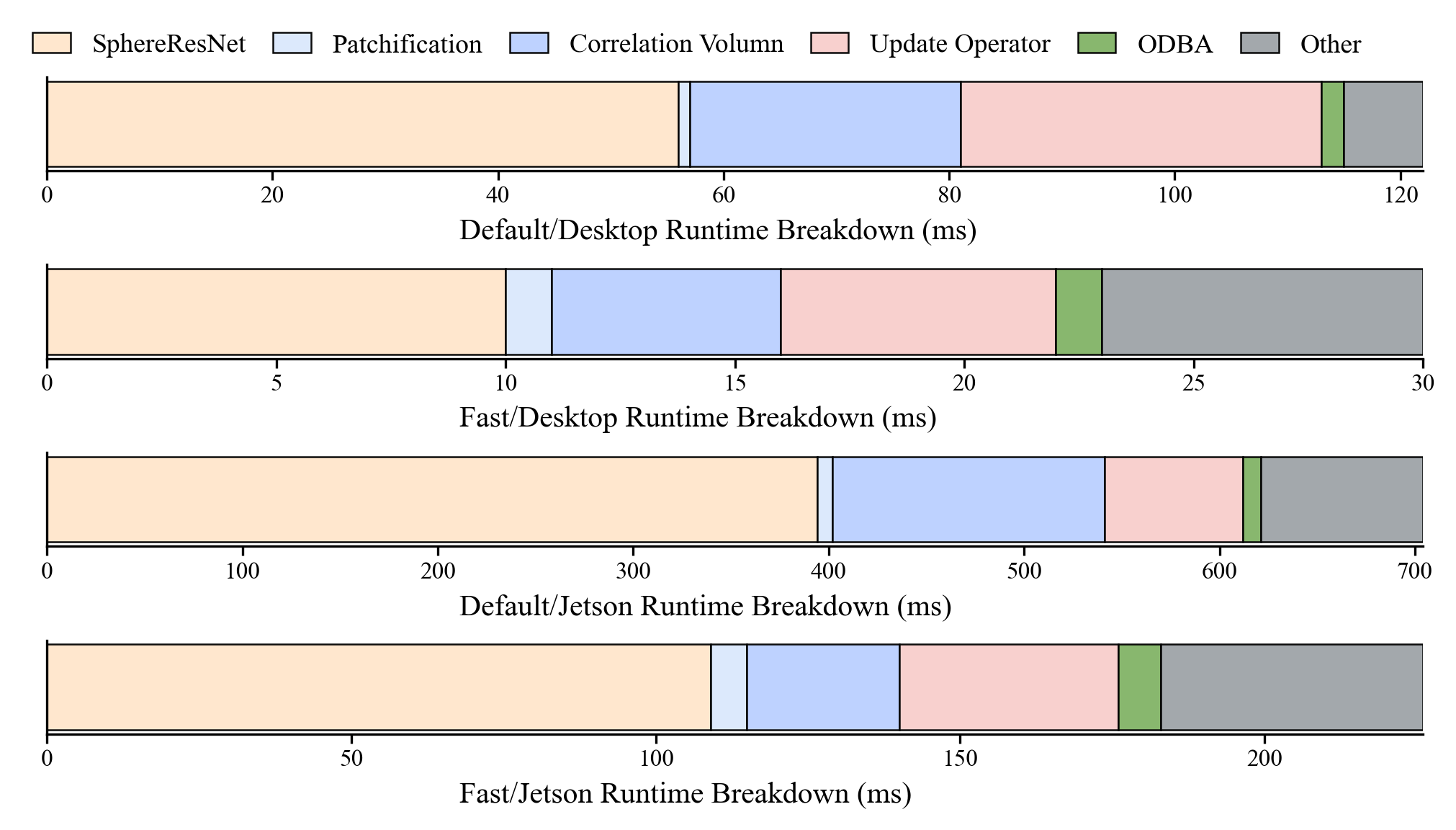}
    \caption{
    \changed{Per-module runtime decomposition across settings and deployment environments. }
    }
    \vspace{-0.5cm}
    \label{fig:runtime}
\end{figure}

To assess the applicability of 360DVO in robotics, we compare accuracy (ATE-RMSE) and throughput (FPS) for 360DVO and OpenVSLAM~\cite{openvslam} on a Jetson Orin Developer Kit. As shown in Tab.~\ref{tab:jetson}, 360DVO (default) is the most accurate but also the slowest. 360DVO (fast) recovers efficiency ($\approx5$ FPS) while remaining competitive in accuracy, outperforming OpenVSLAM (5.07 vs. 6.38).
Although computationally slightly more expensive than conventional OVO methods, 360DVO achieves noticeably higher accuracy and more robust tracking. We believe this substantial performance gain offsets the increased computational cost.
A quantitative breakdown of per‑module runtime for 360DVO under different settings and environments is shown in Fig.~\ref{fig:runtime}. Modules operating on high‑resolution feature maps constitute the principal bottleneck. 
While reducing input resolution or increasing hardware compute capacity can improve inference speed, developing a more lightweight yet effective feature extractor is necessary for enabling real-time operation on embedded platforms, which is a promising direction for future work.

\vspace{-0.2cm}
\section{Conclusion}\label{sec:conclusion}
We present 360DVO, the first deep learning–based omnidirectional visual odometry (OVO) system. 360DVO extracts sparse, informative patch features via a distortion-aware spherical feature extractor, while jointly optimizing camera poses and 3D points through an omnidirectional differentiable bundle adjustment module. To enable comprehensive OVO evaluation, we also introduce a new real-world benchmark dataset. Extensive experiments demonstrate that 360DVO achieves state-of-the-art performance, substantially outperforming prior methods across accuracy and robustness metrics.

\noindent{\textbf{Acknowledgement.}} The authors would like to express their sincere gratitude to the “Sustainable Smart Campus as a Living Lab” (SSC) program at HKUST and ePropulsion for their support.

\bibliographystyle{IEEEtran} 
\bibliography{reference}

\clearpage
\section*{\Large Appendix}
\vspace{0.5cm}

\section{Jacobians of ODBA}
The objective of ODBA is to minimize the cost function $E$: 
\begin{equation}
\underset{\mathbf{T}_i,\mathbf{T}_j,\mathbf{d}}{\textrm{arg\,min}}\enspace E\,\,=\sum_{(k,j)\in\mathcal{E}} \left\| \mathbf{p}^\star_{kj} - \mathbf{p}_{kj}' \right\|^2_{\sum_{kj}}.
\end{equation}

To obtain the Hessian matrix, we need to compute the Jacobian of the reprojection patch $\mathbf{p}'$ with respect to $\mathbf{T}_i$, $\mathbf{T}_j$, and $\mathbf{d}$. Since $\mathbf{T}_i,\mathbf{T}_j\in \textbf{SE}(3)$, we use the exponential mapping to transform them into the tangent plane, specifically letting $\mathbf{T}_n:=e^{\Delta\xi_n}\mathbf{T}_n$, $n\in\{i,j\}$. Then we represent $\mathbf{T}_i$ and $\mathbf{T}_j$ using 6-dimensional vectors 
$\xi_i,\xi_j\in \mathfrak{se} (3)$. After local parametrization, we compute the Jacobians using the chain rule: \begin{equation}
    J_j=\frac{\partial\mathbf{p}'}{\partial\xi_j}=\frac{\partial\Pi(\mathbf{X}')}{\partial\mathbf{X}'}\frac{\partial\mathbf{X}'}{\partial\xi_j},\quad J_i=-J_j\cdot\mathrm{Adj}_{\mathbf{T}_{ij}},
\end{equation}
\begin{equation}
    J_d=\frac{\partial\mathbf{p}'}{\partial d}=\frac{\partial\Pi(\mathbf{X}')}{\partial\mathbf{X}'}\frac{\partial\mathbf{X}'}{\partial\mathbf{X}}\frac{\partial\Pi^{-1}(\mathbf{p},d)}{\partial d}=\frac{\partial\Pi(\mathbf{X}')}{\partial\mathbf{X}'}\cdot\mathbf{T}_{ij}\cdot\frac{\partial\Pi^{-1}(\mathbf{p},d)}{\partial d}~.
\end{equation}

For reference, we restate the projection function $\Pi$ and inverse projection function $\Pi^{-1}$ of spherical camera model: 
\begin{equation}\label{eq:spherical_project}
    \Pi(\mathbf{X})=\begin{bmatrix}
        u\\v\\1
    \end{bmatrix}=\mathbf{K}\begin{bmatrix}
        \theta\\\phi\\1
    \end{bmatrix}=\mathbf{K}\begin{bmatrix}
        \arctan(x/z)\\-\arcsin(d\cdot y)\\1
    \end{bmatrix},
\end{equation}
\begin{equation}\label{eq:spherical_project_inv}    
\Pi^{-1}(\mathbf{p},\mathbf{d})=\begin{bmatrix}
        x\\y\\z\\1
    \end{bmatrix}=\frac{1}{\mathbf{d}}\begin{bmatrix}
        \cos(\phi)\sin(\theta)\\-\sin(\phi)\\\cos(\phi)\cos(\theta)\\\mathbf{d}
    \end{bmatrix},
\end{equation}
\begin{equation*}
    \mathbf{K}
    =\begin{bmatrix}
        W/2\pi&0&W/2\\0&-H/\pi&H/2\\0&0&1
    \end{bmatrix},\quad
    \begin{bmatrix}
        \theta\\\phi\\1
    \end{bmatrix}=\mathbf{K}^{-1}\begin{bmatrix}
        u\\v\\1
    \end{bmatrix},
\end{equation*}
where $d=1/\sqrt{x^2+y^2+z^2}$ represents the inverse distance from the 3D point to the center of the unit sphere.
Letting $\hat{d}=1/\sqrt{x^2+z^2}$, the partial derivative of the projection and the inverse projection functions are derived from Eq.~\ref{eq:spherical_project} and Eq.~\ref{eq:spherical_project_inv} as: 
\begin{equation}
    \frac{\partial\Pi(\mathbf{X})}{\partial\mathbf{X}}=\begin{bmatrix}
         \hat{d}^2W/2\pi&0\\0&-d^2H/\pi
    \end{bmatrix}
    \begin{bmatrix}
        z&0&-x&0\\
        xy\hat{d}&1/\hat{d}&yz\hat{d}&0
    \end{bmatrix},
\end{equation}
\begin{equation}
\frac{\partial\Pi^{-1}(\mathbf{p},\mathbf{d})}{\partial \mathbf{d}}=-\frac{1}{\mathbf{d}^2}
    \begin{bmatrix}
       \cos(\phi)\sin(\theta)\\-\sin(\phi)\\\cos(\phi)\cos(\theta)\\0
    \end{bmatrix}.
\end{equation}
Using the local parametrization and adjoint operator, the 3D point transformation can be expressed as:
\begin{align}\label{3d point transformation}    \mathbf{X}'&=e^{\Delta\xi_j}\mathbf{T}_j\cdot(e^{\xi_i}\mathbf{T}_i)^{-1}\cdot\mathbf{X}
\\&=e^{\Delta\xi_j}\cdot e^{-\mathrm{Adj}_{\mathbf{T}_{ij}}\Delta\xi_i}\cdot\mathbf{T}_{ij}\cdot\mathbf{X},
\end{align} 
where $\mathbf{X}'$ denotes the 3D point after transformation. The partial derivatives of $\mathbf{X}'$ with respect to $\xi_i$ and $\xi_j$ are computed as
\begin{equation}\label{Jacobian_1}
    \frac{\partial\mathbf{X}'}{\partial\xi_j}=\begin{bmatrix}1&0&0&0&z'&-y'\\0&1&0&-z'&0&x'\\0&0&1&y'&-x'&0\\0&0&0&0&0&0\end{bmatrix}, 
\end{equation}
\begin{equation}    
\frac{\partial\mathbf{X}'}{\partial\xi_i}=-\frac{\partial\mathbf{X}'}{\partial\xi_j}\cdot \mathrm{Adj}_{\mathbf{T}_{ij}}.
\end{equation}

After computing the partial derivatives, we assemble the Jacobians $J_i,J_j\in\mathbb{R}^{2\times6}$ and $J_d\in\mathbb{R}^{2\times1}$.

\section{Datasets Details}

\begin{figure*}[!t]
    \centering
    \includegraphics[width=0.95\linewidth]{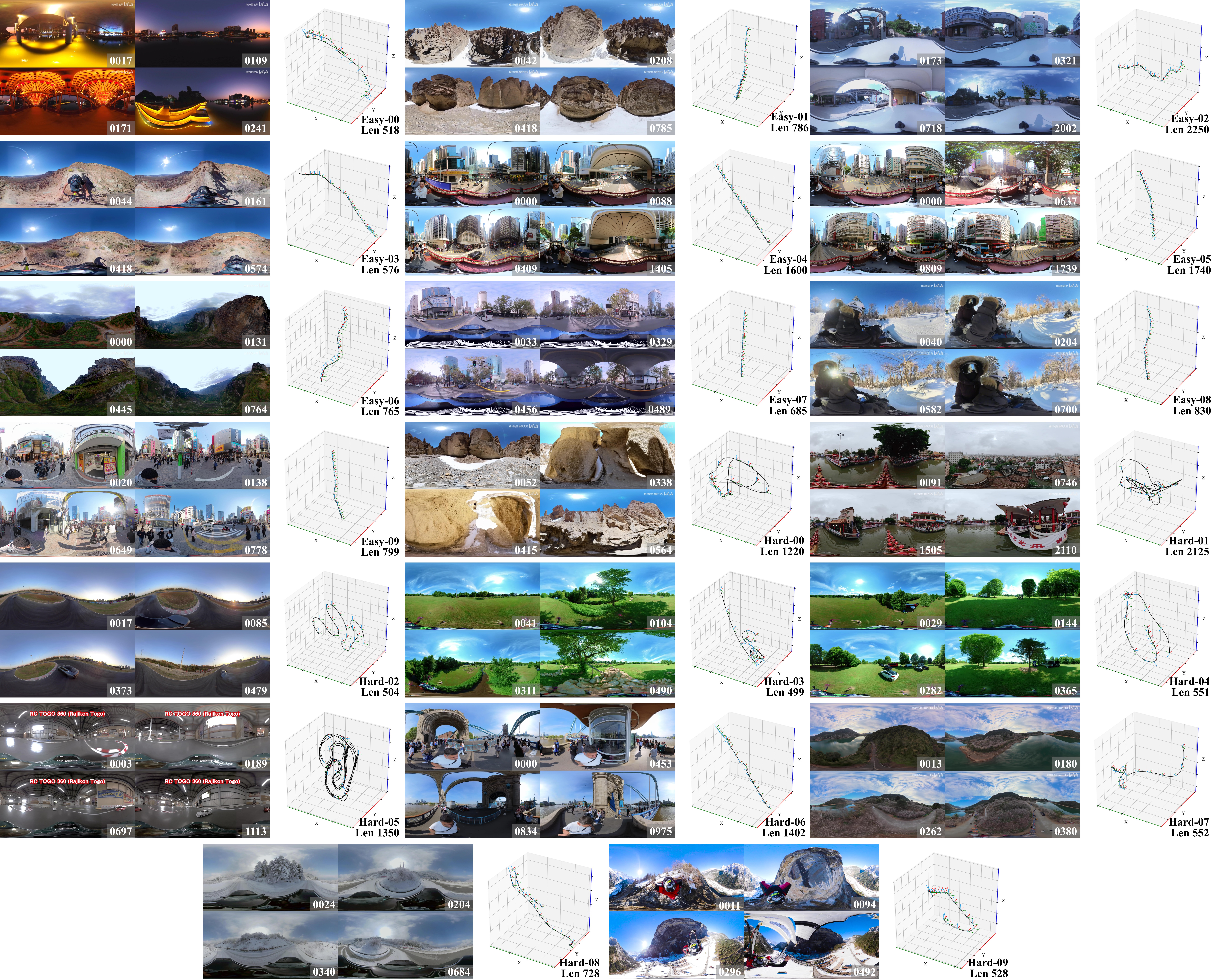}
    \caption{All 20 sequences from the 360DVO dataset, as a complement to Fig.~\ref{fig:dataset} of the main paper.
    }
    \label{fig:appdata}
\end{figure*}

\begin{figure*}[!h]
    \centering
    \includegraphics[width=\linewidth]{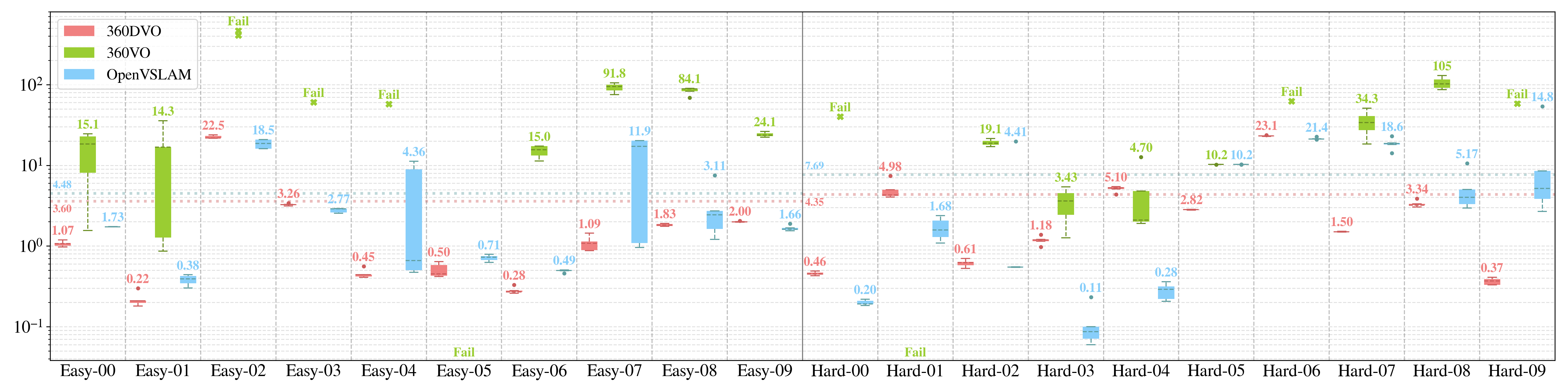}
    \caption{Boxplot results of OVO methods on the 360DVO dataset. Our 360DVO runs stably with lowest variations. }
    \label{fig:box}
\end{figure*}

\begin{table}[h]
    \centering
\resizebox{0.8\linewidth}{!}{
    \begin{tabular}{lc c cccc c}
    \toprule
    \multicolumn{2}{c}{\multirow{2}{*}{Seq.}}&\multirow{2}{*}{Len.}&\multicolumn{4}{c}{Challenges}&\multirow{2}{*}{Source}\\
    \cmidrule{4-7}
    &&&CT&IV&VR&DO&\\
    \midrule
    \multirow{10}{*}{Easy}&00& 518& &\checkmark&& &Bilibili \\
    &01& 786& &&& &Bilibili \\
    &02& 2250& &\checkmark&& &Bilibili \\
    &03& 576& &&\checkmark& &Bilibili \\
    &04& 1600& &\checkmark&&\checkmark &Self-collected \\

    &05& 1740& &\checkmark&&\checkmark &Self-collected \\
    &06& 765& &&\checkmark& &Bilibili \\
    &07& 685& &\checkmark&&\checkmark &Bilibili \\
    &08& 830& &&& &Bilibili \\
    &09& 799& &\checkmark&&\checkmark &Bilibili \\

    \midrule
    
    \multirow{10}{*}{Hard}&00& 1220& \checkmark&&\checkmark& &Bilibili \\
    &01& 2125& \checkmark&&\checkmark& &Bilibili \\
    &02& 504& \checkmark&&\checkmark& &Bilibili \\
    &03& 499& \checkmark&&\checkmark& &360VOTS~\cite{360VOTS} \\
    &04& 551& \checkmark&&\checkmark& &360VOTS~\cite{360VOTS} \\

    &05& 1350& \checkmark&&\checkmark& &360VOTS~\cite{360VOTS} \\
    &06& 1402& \checkmark&&&\checkmark &Bilibili \\
    &07& 552& \checkmark&&\checkmark& &Bilibili \\
    &08& 728& \checkmark&\checkmark&& &Bilibili \\
    &09& 528& \checkmark&\checkmark&\checkmark&\checkmark &Bilibili \\
    
    \bottomrule
    \end{tabular}
}
    \caption{Dataset details, including sequence length, challenges, and sources. The challenges are specified by Complex Trajectory (CT), Illumination Variations(IV), Violent Rotation(VR), and Dynamic Objects (DO).}
    \label{tab:my_label}
\end{table}

We provide illustrations for all 20 sequences and plot the 3D trajectories in Fig.~\ref{fig:appdata}. We evaluate the sequences based on multiple metrics and classify them according to the complexity of the trajectories. We report the sequence length, video sources, and challenge metrics, including Complex Trajectory, Illumination Variations, Violent Rotation and Dynamic Objects in Tab.~\ref{tab:my_label}.

\section{Additional Experiment Results}\label{sec::additional_exp}

\noindent{\textbf{Experiment Details. } In the experiments, we define results with track loss exceeding half of the total frames as failed results. For results with misaligned timestamps, we only compare existing trajectories. Therefore, some of the plotted experimental trajectories may not represent complete paths. All methods are run 5 times under the same configuration and environment. 
}

\noindent{\textbf{Boxpolt Results. }} We provide a boxplot of the experimental results, shown in Fig.~\ref{fig:box}. We report the results of five runs for all OVO methods on our real-world dataset. 360DVO achieves the lowest error and variance among the OVO methods. With the ability to learn distortion-free features through SphereResNet and optimize camera poses and depths using the ODBA component, 360DVO can reliably and accurately estimate trajectories in challenging environments. The direct method-based 360VO~\cite{360VO} fails on nearly half of the sequences. Although OpenVSLAM~\cite{openvslam} has high success and accuracy rates, its performance remains unstable on some sequences, exhibiting significant variance. 

\begin{figure*}[h]
    \centering
    \includegraphics[width=\linewidth]{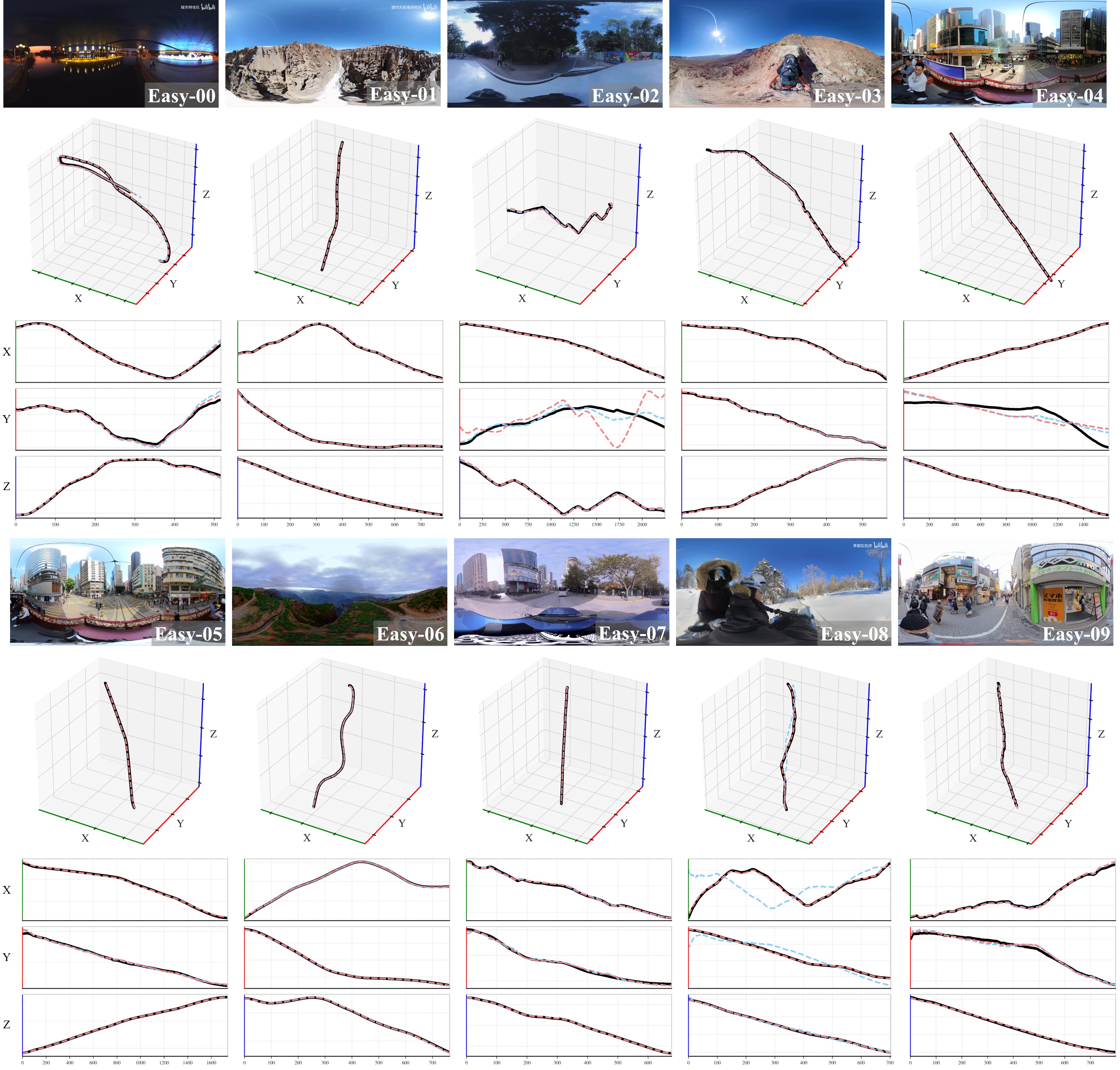}
    \caption{Visual comparison of the resulting trajectories in \textit{Easy} sequences, as a complement to Fig.~\ref{fig:res} of the main paper.. The ground truth, results of 360DVO, and results of OpenVSLAM~\cite{openvslam} are marked in \textbf{black solid lines}, \textbf{\dovored{red dashed lines}}, and \textbf{\openvblue{blue dashed lines} } separately. 
    For each sequence, we compare the overall shapes of their trajectories in the 3D space while examining the variations across all frames on each of the X, Y, and Z axes.
    }
    \label{fig:appreseasy}
\end{figure*}

\begin{figure*}[h]
    \centering
    \includegraphics[width=\linewidth]{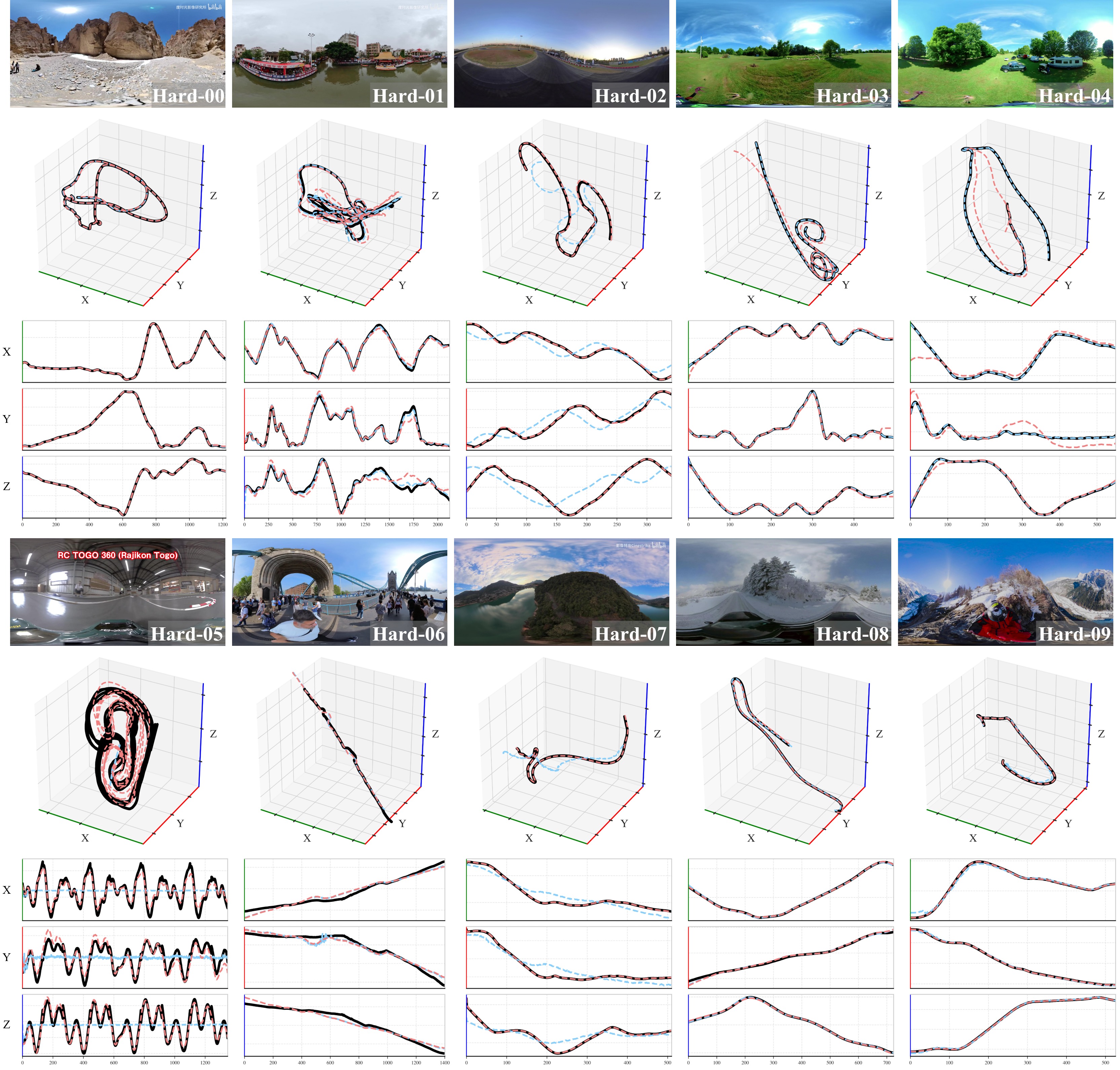}
    \caption{Visual comparison of the resulting trajectories in \textit{Hard} sequences, as a complement to Fig.~\ref{fig:res} of the main paper.. The ground truth, results of 360DVO, and results of OpenVSLAM~\cite{openvslam} are marked in \textbf{black solid lines}, \textbf{\dovored{red dashed lines}}, and \textbf{\openvblue{blue dashed lines} } separately. 
    For each sequence, we compare the overall shapes of their trajectories in the 3D space while examining the variations across all frames on each of the X, Y, and Z axes.
    }
    \label{fig:appreshard}
\end{figure*}

\noindent{\textbf{Multidimensional Comparison of Trajectories. }} 
In Fig.~\ref{fig:appreseasy} and Fig.~\ref{fig:appreshard}, we provide a multidimensional visualization of the trajectories for OpenVSLAMand 360DVO across all sequences. The results of the two methods are compared against the ground truth trajectories in 3D, as well as on the x-axis, y-axis, and z-axis. Multidimensional visual comparisons clearly show that 360DVO closely aligns with the ground truth trajectories in most sequences, significantly outperforming OpenVSLAM.

\end{document}